\documentclass{article}

\PassOptionsToPackage{numbers}{natbib}

\usepackage[final]{neurips_2022}

\usepackage[utf8]{inputenc} %
\usepackage[T1]{fontenc}    %
\usepackage{hyperref}       %
\usepackage{url}            %
\usepackage{booktabs}       %
\usepackage{amsmath}        %
\usepackage{amssymb}        %
\usepackage{amsfonts}       %
\usepackage{bm}             %
\usepackage{xfrac}          %
\usepackage{microtype}      %
\usepackage{xcolor}         %
\usepackage{graphicx}       %
\usepackage{enumitem}       %
\usepackage{multirow}       %
\usepackage{wrapfig}        %
\usepackage{subcaption}     %
\usepackage{float}          %
\usepackage{lscape}         %
\usepackage{algorithm}      %
\usepackage[noend]{algorithmic}
\usepackage[para]{footmisc} %

\newcommand{\adj}{\mathbf{A}}
\newcommand{\feat}{\mathbf{X}}
\newcommand{\pertmodel}{\Phi}
\newcommand{\lab}{\mathbf{y}}
\newcommand{\hid}{\mathbf{H}}
\newcommand{\weight}{\mathbf{W}}
\newcommand{\bias}{\mathbf{b}}
\newcommand{\pertadj}{\tilde{\adj}}
\newcommand{\diffadj}{\delta\adj}
\newcommand{\params}{\theta}
\newcommand{\model}{f_{\params^*}}
\newcommand{\loss}{\ell}
\DeclareMathOperator{\relu}{relu}
\DeclareMathOperator{\elu}{elu}
\DeclareMathOperator{\dropout}{dropout}

\graphicspath{{./assets/}}

\title{Are Defenses for Graph Neural Networks Robust?}

\author{%
  Felix Mujkanovic\textsuperscript{\ensuremath{1}}\thanks{equal contribution}\,\,, Simon Geisler\textsuperscript{\ensuremath{1}}$^{\fnsymbol{footnote}}$, Stephan Günnemann\textsuperscript{\ensuremath{1}}, Aleksandar Bojchevski\textsuperscript{\ensuremath{2}} \\
  \textsuperscript{\ensuremath{1}}{Dept. of Computer Science \& Munich Data Science Institute, Technical University of Munich}\\
  \textsuperscript{\ensuremath{2}}{CISPA Helmholtz Center for Information Security}\\
  \texttt{\{f.mujkanovic, s.geisler, s.guennemann\}@tum.de | bojchevski@cispa.de}
}

\begin{document}

\maketitle

\begin{abstract}
    A cursory reading of the literature suggests that we have made a lot of progress in designing effective adversarial defenses for Graph Neural Networks (GNNs). Yet, the standard methodology has a serious flaw -- virtually all of the defenses are evaluated against non-adaptive attacks leading to overly optimistic robustness estimates. We perform a thorough robustness analysis of 7 of the most popular defenses spanning the entire spectrum of strategies, i.e., aimed at improving the graph, the architecture, or the training. The results are sobering -- most defenses show no or only marginal improvement compared to an undefended baseline. We advocate using custom adaptive attacks as a gold standard and we outline the lessons we learned from successfully designing such attacks. Moreover, our diverse collection of perturbed graphs forms a (black-box) unit test offering a first glance at a model's robustness.\looseness=-1\footnote[1]{Project page: \resizebox{0.885\width}{\height}{\url{https://www.cs.cit.tum.de/daml/are-gnn-defenses-robust/}}}
\end{abstract}

\section{Introduction}\label{sec:introduction}

The vision community learned a bitter lesson -- we need specific carefully crafted attacks to properly evaluate the adversarial robustness of a defense. Consequently, adaptive attacks are considered the gold standard~\citep{tramer_adaptive_2020}. This was not always the case; until recently, most defenses were tested only against relatively weak static attacks. The turning point was \citet{carlini_adversarial_2017}'s work showing that 10 methods for detecting adversarial attacks can be easily circumvented. Shortly after, \citet{athalye_obfuscated_2018} showed that 7 out of the 9 defenses they studied can be broken since they (implicitly) rely on obfuscated gradients. So far, this bitter lesson is completely ignored in the graph domain.

\begin{figure}[H]
    \centering
    \captionsetup[subfigure]{justification=centering}
    {
        \captionsetup[subfigure]{margin={1.4cm,0cm}}
        \subcaptionbox{Global, Poisoning}{\includegraphics[scale=0.64]{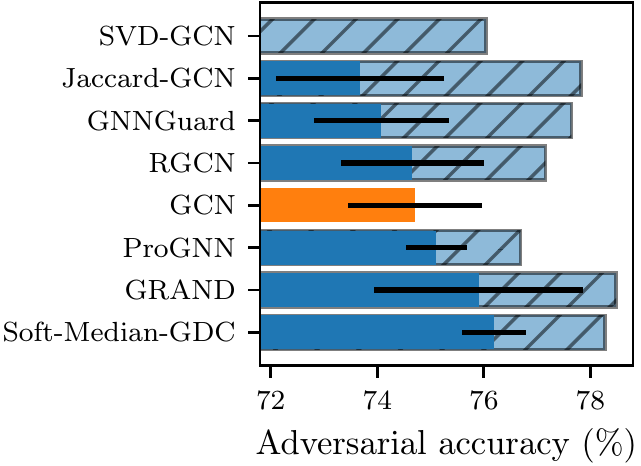}}
    }%
    \hfill
    \subcaptionbox{Global, Evasion}{\includegraphics[scale=0.64]{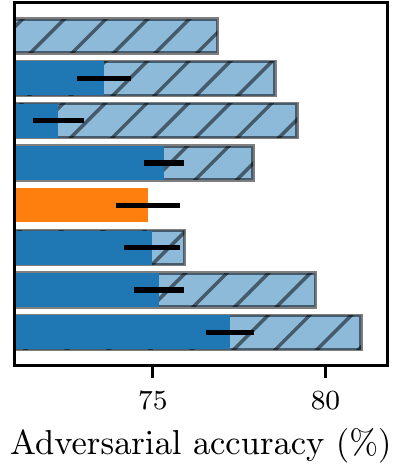}}
    \hfill
    \subcaptionbox{Local, Poisoning}{\includegraphics[scale=0.64]{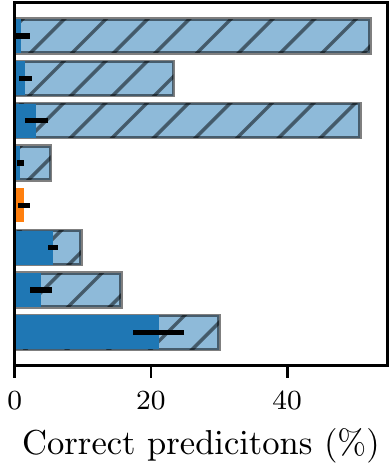}}
    \hfill
    {
        \captionsetup[subfigure]{margin={0cm,1.6cm}}
        \subcaptionbox{Local, Evasion}{\includegraphics[scale=0.64]{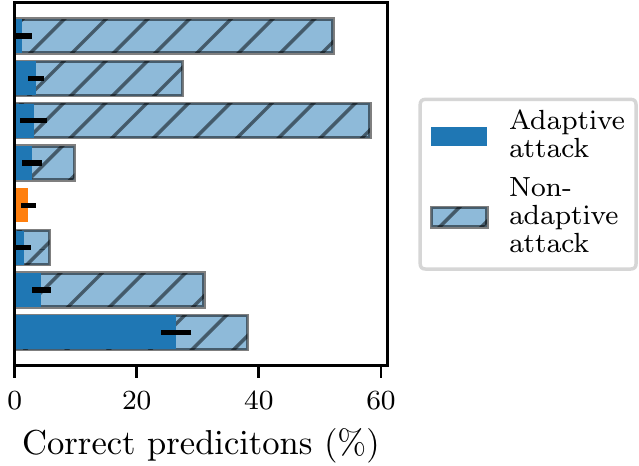}}
    }%
    \caption{Adaptive attacks draw a different picture of robustness. All defenses are less robust than reported, with an undefended GCN~\citep{kipf_semi-supervised_2017} outperforming some. We show results on Cora~ML for both poisoning (attack before training) and evasion (attack after training), and both global (attack the test set jointly) and local (attack individual nodes) setting. The perturbation budget is relative w.r.t. the \#edges for global attacks (5\% evasion, 2.5\% poisoning) and w.r.t. the degree for local attacks (100\%). In (a)/(b) SVD-GCN is catastrophically broken -- our adaptive attacks reach 24\%/9\% (not visible). Note that our non-adaptive attacks are already stronger than what is typically used (see \autoref{sec:adaptive_attacks}).}
    \label{fig:superstar}
\end{figure}

Virtually no existing work that proposes an allegedly robust Graph Neural Network (GNN) evaluates against adaptive attacks, leading to overly optimistic robustness estimates. To show the seriousness of this methodological flaw we categorize 49 works that propose a robust GNN and are published at major conferences/journals. We then choose one defense per category (usually the most highly cited). Not surprisingly, we show that none of the assessed models are as robust as originally advertised in their respective papers. In \autoref{fig:superstar} we summarize the results for 7 of the most popular defenses, spanning the entire spectrum of strategies (i.e., aimed at improving the graph, the architecture, or the training, see \autoref{tab:categorization}).

We see that in both local and global settings, as well as for both evasion and poisoning, the adversarial accuracy under our adaptive attacks is significantly smaller compared to the routinely used non-adaptive attacks. Even more troubling is that many of the defenses perform worse than an undefended baseline (a vanilla GCN \citep{kipf_semi-supervised_2017}). Importantly, the 7 defenses are not cherry-picked. We report the results for each defense we assessed and selected each defence before running any experiments.

Adversarial robustness measures the local generalization capabilities of a model, i.e., sensitivity to (bounded) worst-case perturbations. Certificates typically provide a lower bound on the actual robustness while attacks provide an upper bound. Since stronger attacks directly translate into tighter bounds our goal is to design the strongest attack possible. Our adaptive attacks have perfect knowledge of the model, the parameters, and the data, including all defensive measures. In contrast, non-adaptive attacks (e.g., transferred from an undefended proxy or an attack lacking knowledge about defense measures) only show how good the defense is at suppressing a narrow subset of input perturbations.\footnote{From a security perspective non-adaptive attacks (typically transfer attacks) are also relevant since a real-world adversary is unlikely to know everything about the model and the data.}

\citet{tramer_adaptive_2020} showed that even adaptive attacks can be tricky to design with many subtle challenges. The graph domain comes with additional challenges since graphs are typically sparse and discrete and the representation of any node depends on its neighborhood. For this reason, we describe the recurring themes, the lessons learned, and our systematic methodology for designing strong adaptive attacks for all examined models. Additionally, we find that defenses are \emph{sometimes} sensitive to a common attack vector and transferring attacks can also be successful. Thus, the diverse collection of perturbed adjacency matrices resulting from our attacks forms a (black-box) unit test that any truly robust model should pass before moving on to adaptive evaluation. In summary:
\begin{itemize}[left=5pt, noitemsep]
    \item We survey and categorize \emph{49 defenses} published across prestigious machine learning venues.
    \item We design custom attacks for 7 defenses (14\%), covering the spectrum of defense techniques. All examined models forfeit a large fraction of previously reported robustness gains.
    \item We provide a transparent methodology and guidelines for designing strong adaptive attacks.
    \item Our collection of perturbed graphs can serve as a robustness unit test for GNNs.
\end{itemize}

\section{Background and preliminaries}\label{sec:preliminaries}

We follow the most common setup and assume GNN~\citep{gilmer_neural_2017, kipf_semi-supervised_2017} classifiers \(f_{\params}(\adj, \feat)\) that operate on a symmetric binary adjacency matrix \(\adj \in \{0, 1\}^{n \times n}\) with binary node features \(\feat \in \{0, 1\}^{n \times d}\) and node labels \(\lab \in \{1, 2, \dots, C\}^{n}\) where \(C\) is the number of classes, \(n\) is the number of nodes, and \(m\) the number of edges. A poisoning attack perturbs the graph (flips edges) prior to training, optimizing
\begin{equation}\label{eq:poisoning}
    \max_{\pertadj \in \pertmodel(\adj)} \loss_{\text{attack}}(\model(\pertadj, \feat), \lab) \text{\quad s.t. \quad} \params^* = \arg\min_{\params} \loss_{\text{train}}(f_{\params}(\pertadj, \feat), \lab)
\end{equation}
where \(\loss_{\text{attack}}\) is the attacker's loss, which is possibly different from \(\loss_{\text{train}}\) (see \autoref{sec:methodology}). In an evasion attack, \(\params^*\) is kept fixed and obtained by training on the clean graph \(\min_{\params} \loss_{\text{train}}(f_{\params}(\adj, \feat), \lab)\). In both cases, the locality constraint \(\pertmodel(\adj)\) enforces a budget \(\Delta\) by limiting the perturbation to an \(L_0\)-ball around the clean adjacency matrix: \(\|\pertadj - \adj\|_0 \le 2 \Delta\). Attacks on \(\feat\) also exist, however, this scenario is not considered by the vast majority of defenses. For example, only one out of the seven examined ones also discusses feature perturbations. We refer to \autoref{sec:appendix:feature_perturbations} for more details on adaptive feature attacks.

\textbf{Threat model.}
Our attacks aim to either cause misclassification of the entire test set (\emph{global}) or a single node (\emph{local}). To obtain the strongest attack possible (i.e., tightest robustness upper bound), we use white-box attacks. We do not constrain the attacker beyond a simple budget constraint that enforces a maximum number of perturbed edges. For our considerations on unnoticeability, see \autoref{sec:appendix:attack_overview}.

\textbf{Greedy attacks.}
Attacking a GNN typically corresponds to solving a constrained discrete non-convex optimization problem that -- evident by this work -- is hard to solve. Commonly, approximate algorithms are used to to tackle these optimization problems. For example, the single-step Fast Gradient Attack (FGA) flips the edges whose gradient (i.e., \(\nabla_{\adj} \loss_{\text{train}} (\model(\adj, \feat), \lab)\)) most strongly indicates so. On the other hand, Nettack~\citep{zugner_adversarial_2018} and Metattack~\citep{zugner_adversarial_2019} are greedy multi-step attacks. The greedy approaches have the nice side-effect that an attack for a high budget \(\Delta\) directly gives all attacks for budgets lower than \(\Delta\). On the other hand, they tend to be relatively weaker.

\textbf{Projected Gradient Descent (PGD).}
Alternatively, PGD~\citep{xu_topology_2019} has been applied to GNNs where the discrete adjacency matrix is relaxed to \([0, 1]^{n \times n}\) during the gradient-based optimization and the resulting weighted change reflects the probability of flipping an edge. After each gradient update, the changes are projected back such that the budget holds in expectation \(\|\mathbb{E}[\pertadj] - \adj\|_0 \le 2\Delta\). Finally, multiple samples are obtained and the strongest perturbation \(\pertadj\) is chosen that obeys the budget \(\Delta\). The biggest caveats while applying \(L_0\)-PGD are the relaxation gap and limited scalability (see \citet{geisler_robustness_2021} for a detailed discussion and a scalable alternative).

\textbf{Evasion vs. poisoning.}
Evasion can be considered the easier setting from an attack perspective since the model is fixed \(\model\). For poisoning, on the other hand, the adjacency matrix is perturbed before training (\autoref{eq:poisoning}). Two general strategies exist for poisoning attacks: (1) transfer a perturbed adjacency matrix from an evasion attack~\citep{zugner_adversarial_2018}; or (2) attack directly by, e.g., unrolling the training procedure to obtain gradients through training~\citep{zugner_adversarial_2019}. \citet{xu_topology_2019} propose to solve~\autoref{eq:poisoning} with alternating optimization which was shown to be even weaker than the evasion transfer (1). Note that evasion is particularly of interest for inductive learning and poisoning for transductive learning.

\section{Adversarial defenses}\label{sec:defenses}

We select the defenses s.t. we capture the entire spectrum of methods improving robustness against structure perturbations. For the selection, we extend the taxonomy proposed in~\citep{GNNBook-ch8-gunnemann}. We selected the subset without cherry-picking based on the criteria elaborated below before experimentation.

\textbf{Taxonomy.}
The top-level categories are \emph{improving the graph} (e.g., preprocessing), \emph{improving the training} (e.g., adversarial training or augmentations), and \emph{improving the architecture}. Many defenses for structure perturbations either fall into the category of improving the graph or adaptively weighting down edges through an improved architecture. Thus, we introduce further subcategories. Similar to \citep{GNNBook-ch8-gunnemann}'s discussion, unsupervised improvement of the graph finds clues in the node features and graph structure, while supervised improvement incorporates gradient information from the learning objective. Conversely, for adaptive edge weighting, we identify three prevalent approaches: rule-based (e.g., using a simple metric), probabilistic (e.g., modeling a latent distribution), and robust aggregations (e.g., with guarantees). We assign each defense to the most fitting taxon (details in \autoref{sec:appendix:taxonomy}).

\textbf{Selected defenses.}
To evaluate a diverse set of defenses, we select one per leaf taxon.\footnote{The only exception is unsupervised graph improvement, as it contains two of the most popular approaches, which rely on orthogonal principles. One filters edges based on the node features~\citep{wu_adversarial_2019}, the other uses a low-rank approximation of the adjacency matrix~\citep{entezari_all_2020}.} We prioritize highly cited defenses published at renowned venues with publicly available code. We implement all defenses in one unified pipeline. We present the categorization of defenses and our selection in \autoref{tab:categorization}. Similarly to \citet{tramer_adaptive_2020}, we exclude defenses in the ``robust training'' category (see \autoref{sec:appendix:adversarial_training} for a discussion). Two of the three models in the ``miscellaneous'' category report some improvement in robustness, but they are not explicitly designed for defense purposes so we exclude them from our study. Some works evaluate only against evasion \citep{wu_adversarial_2019}, others only poisoning \citep{entezari_all_2020, feng_graph_2021, zhang_gnnguard_2020}, and the rest tackle both \citep{geisler_robustness_2021, jin_graph_2020, zhu_robust_2019}. In some cases the evaluation setting is not explicitly stated and inferred by us. For completeness, we consider each defense in all four settings (local/global and evasion/poisoning). Next, we provide a short summary of the key ideas behind each defense (details in \autoref{sec:appendix:defenses}).

\begin{table}[ht]
    \centering
    \caption{Categorization of selected defenses. Our taxonomy extends the one by \citet{GNNBook-ch8-gunnemann}.}
    \label{tab:categorization}
    \vspace{0.5em}
    \resizebox{0.95\linewidth}{!}{
        \begin{tabular}{lll|ll}
            \toprule
            \multicolumn{3}{c|}{Taxonomy} & Selected Defenses & Other Defenses \\
            \midrule \midrule
            \multirow{2}[2]{*}{\begin{tabular}[c]{@{}l@{}}Improving \\ graph\end{tabular}} & \multicolumn{2}{l|}{Unsupervised} & \begin{tabular}[c]{@{}l@{}}Jaccard-GCN~\citep{wu_adversarial_2019} \\ SVD-GCN~\citep{entezari_all_2020}\end{tabular} & \citep{duan_aane_2020, ioannidis_unveiling_2021, xiao_lightweight_2021, zhang_comparing_2019, zhang_detection_2021} \\
            \cmidrule{2-5}
            & \multicolumn{2}{l|}{Supervised} & ProGNN \citep{jin_graph_2020} & \citep{xu_speedup_2021, tao_adversarial_2021, zhang_graph_2022} \\
            \midrule
            \multirow{2}[2]{*}{\begin{tabular}[c]{@{}l@{}}Improving \\ training\end{tabular}} & \multicolumn{2}{l|}{Robust training} & n/a (see \autoref{sec:appendix:adversarial_training}) & \citep{chen_smoothing_2020, deng_batch_2019, feng_graph_2021-1, hu_robust_2021, jin_latent_2019, hutter_robust_2021, sun_virtual_2019, xu_unsupervised_2022, xu_topology_2019, xu_towards_2020} \\
            \cmidrule{2-5}
            & \multicolumn{2}{l|}{Further training principles} & GRAND \citep{feng_graph_2021} & \citep{chang_not_2021, elinas_variational_2020, jin_power_2021, regol_node_2022, tang_transferring_2020, you_graph_2020, zheng_robust_2020, zhuang_defending_2022, zhuang_how_2022} \\
            \midrule
            \multirow{4}[4]{*}{\begin{tabular}[c]{@{}l@{}}Improving \\ architecture\end{tabular}} & \multirow{3}[3]{*}{\begin{tabular}[c]{@{}l@{}}Adaptively \\ weighting \\ edges\end{tabular}} & Rule-based & GNNGuard \citep{zhang_gnnguard_2020} & \citep{jin_node_2021, liu_graph_2021, liu_elastic_2021, zhang_feature-importance-aware_2020} \\
            \cmidrule{3-5}
            & & Probabilistic & RGCN \citep{zhu_robust_2019} & \citep{chen_enhancing_2021, feng_uncertainty-aware_2021, ioannidis_edge_2020, ioannidis_tensor_2020, luo_learning_2021} \\
            \cmidrule{3-5}
            & & Robust agg. & Soft-Median-GDC \citep{geisler_robustness_2021} & \citep{chen_understanding_2021, geisler_reliable_2020, wang_provably_2020} \\
            \cmidrule{2-5}
            & \multicolumn{2}{l|}{Miscellaneous} & n/a (see above) & \citep{shanthamallu_uncertainty-matching_2021, wang_graph_2020, wu_graph_2020} \\
            \bottomrule
        \end{tabular}
    }
\end{table}

\textbf{Improving the graph.}
The feature-based \emph{Jaccard-GCN} \citep{wu_adversarial_2019} uses a preprocessing step to remove all edges between nodes whose features exhibit a Jaccard similarity below a certain threshold. This was motivated by the homophily assumption which is violated by prior attacks that tend to insert edges between dissimilar nodes. The structure-based \emph{SVD-GCN} \citep{entezari_all_2020} replaces the adjacency matrix with a low-rank approximation prior to plugging it into a regular GNN. This defense was motivated by the observation that the perturbations from Nettack tend to disproportionately affect the high-frequency spectrum of the adjacency matrix. The key idea in \emph{ProGNN} \citep{jin_graph_2020} is to learn the graph structure by alternatingly optimizing the parameters of the GNN and the adjacency matrix (the edge weights). The loss for the latter includes the standard cross-entropy loss, the distance to the original graph, and three other objectives designed to promote sparsity, low rank, and feature smoothness.

\textbf{Improving the training.}
\emph{GRAND} \citep{feng_graph_2021} relies on random feature augmentations (zeroing features) coupled with neighbourhood augmentations \(\bar{\feat} = (\adj\feat + \adj\adj\feat + \cdots)\). All randomly augmented copies of \(\bar{\feat}\) are passed through the same MLP that is trained with a consistency regularization loss.

\textbf{Improving the architecture.}
\emph{GNNGuard} \citep{zhang_gnnguard_2020} filters edges in each message passing aggregation via cosine-similarity (smoothed over layers). In the first layer of \emph{RGCN} \citep{zhu_robust_2019} we learn a Gaussian distribution over the feature matrix and the subsequent layers then manipulate this distribution (instead of using point estimates). For the loss we then sample from the resulting distribution. In addition, in each layer, RGCN assigns higher/lower weights to features with low/high variance. \emph{Soft-Median-GDC} \citep{geisler_robustness_2021} replaces the message passing aggregation function in GNNs (typically a weighted mean) with a more robust alternative by relaxing the median using differentiable sorting.

\textbf{Common themes.}
One theme shared by some defenses is to first discover some property that can discriminate clean from adversarial edges (e.g., high vs. low feature similarity), and then propose a strategy based on that property (e.g., filter low similarity edges). Often they analyze the edges from only a single attack such as Nettack \citep{zugner_adversarial_2018}. The obvious pitfall of this strategy is that the attacker can easily adapt by restricting the adversarial search space to edges that will bypass the defense's (implicit) filter. Another theme is to add additional loss terms to promote some robustness objectives. Similarly, the attacker can incorporate the same terms in the attack loss to negate their influence.

\section{Methodology: How to design strong adaptive attacks}\label{sec:methodology}

In this section, we describe our general methodology and the lessons we learned while designing adaptive attacks. We hope these guidelines can serve as a reference for testing new defenses.

\textbf{Step 1 -- Understand how the defense works}
and categorize it. For example, some defenses rely on preprocessing which filters out edges that meet certain criteria (e.g., Jaccard-GCN~\citep{wu_adversarial_2019}). Others introduce additional losses during training (e.g., GRAND \citep{feng_graph_2021}) or change the architecture (e.g., RGCN \citep{zhu_robust_2019}). Different defenses might need different attacks or impose extra requirements on them.

\textbf{Step 2 -- Probe for obvious weaknesses.}
Some examples include: (a) transfer adversarial edges from another (closely related) model (see also \autoref{sec:black_box_attack}); (b) use a gradient-free (black-box) attack. For example, in our local experiments, we use a \emph{Greedy Brute Force} attack: in each step, it considers all possible single edge flips and chooses the one that contributes most to the attack objective (details in \autoref{sec:appendix:attack_overview}).

\textbf{Step 3 -- Launch a gradient-based adaptive attack.}
For rapid prototyping, use a comparably cheap attack such as FGA, and later advance to stronger attacks like PGD. For poisoning, strongly consider meta-gradient-based attacks like Metattack~\citep{zugner_adversarial_2019} that unroll the training procedure, as they almost always outperform just transferring perturbations from evasion. Unsurprisingly, we find that applying PGD~\citep{xu_topology_2019} on the meta gradients often yields even stronger attacks than the greedy Metattack, and we refer to this new attack as \emph{Meta-PGD} (details in \autoref{sec:appendix:attack_overview}).

\textbf{Step 4 -- Address gradient issues.}
Some defenses contain components that are non-differentiable, lead to exploding or vanishing gradients, or obfuscate the gradients~\citep{athalye_obfuscated_2018}. To circumvent these issues, potentially: (a) adjust the defense's hyperparameters to retain numerical stability; (b) replace the offending component with a differentiable or stable counterpart, e.g., substitute the low-rank approximation of SVD-GCN~\citep{entezari_all_2020} with a suitable differentiable alternative; or (c) remove components, e.g., drop the ``hard'' filtering of edges done in the preprocessing of Soft-Median-GDC~\citep{geisler_robustness_2021}. These considerations also include poisoning attacks, where one also needs to pay attention to all components of the training procedure. For example, we ignore the nuclear norm loss term in the training of ProGNN~\citep{jin_graph_2020} to obtain the meta-gradient. Of course, keep the entire defense intact for its final evaluation on the found perturbations.

\textbf{Step 5 -- Adjust the attack loss.}
In previous works, the attack loss is often chosen to be the same as the training loss, i.e., the cross-entropy (CE). This is suboptimal since CE is not \emph{consistent} according to the definition by \citet{tramer_adaptive_2020} -- higher loss values do not indicate a stronger attack. Thus, we use a variant of the consistent Carlini-Wagner loss~\citep{carlini_towards_2017} for \emph{local} attacks, namely the logit margin (LM), i.e., the logit difference between the ground truth class and most-likely non-true class. However, as discussed by \citet{geisler_robustness_2021}, for \emph{global} attacks the mean LM across all target nodes is still suboptimal since it can ``waste'' budget on already misclassified nodes. Their tanh logit margin (TLM) loss resolves this issue. If not indicated otherwise, we either use TLM or the probability margin (PM) loss -- a slight variant of LM that computes the margin after the softmax rather than before.

\textbf{Step 6 -- Tune the attack hyperparameters}
such as the number of PGD steps, the attack learning rate, the optimizer, etc. For example, for Metattack we observed that using the Adam optimizer~\citep{kingma_adam_2015} can weaken the attack and replacing it with SGD can increase the effectiveness.

\textbf{Lessons learned.}
We provide a detailed description of each adaptive attack and the necessary actions to make it as strong as possible in \autoref{sec:appendix:defenses}. Here, we highlight some important recurring challenges that should be kept in mind when designing adaptive attacks. (1) Numerical issues, e.g., due to division by tiny numbers can lead to weak attacks, and we typically resolve them via clamping. (2) In some cases we observed that for PGD attacks it is beneficial to clip the gradients to stabilize the adversarial optimization. (3) For a strong attack it is essential to tune its hyperparameters. (4) Relaxing non-differentiable components and deactivating operations that filter edges/embeddings based on a threshold in order to obtain gradients for every edge is an effective strategy. (5) If the success of evasion-poisoning transfer depends on a fixed random initialization (see \autoref{sec:appendix:random_seed}), it helps to use multiple clean auxiliary models trained with different random seeds for the PGD attack -- in each PGD step we choose one model randomly. (6) Components that make the optimization more difficult but barely help the defense can be safely deactivated. (7) It is sometimes beneficial to control the randomness in the training loop of Meta-PGD. (8) For Meta-PGD it can help to initialize the attack with non-zero perturbations and e.g., use the perturbed graph of a different attack.

\textbf{Example 1 -- SVD-GCN.}
To illustrate the attack process (especially steps 3 and 4) we present a case study of how we construct an adaptive attack against SVD-GCN. Gradient-free attacks like Nettack do not work well here as they waste budget on adversarial edges which are filtered out by the low-rank approximation (LRA). Moreover, to the demise of gradient-based attacks, the gradients of the adjacency matrix are very unstable due to the SVD and thus less useful. Still, we start with a gradient-based attack as it is easier to adapt, specifically FGA, whose quick runtime enables rapid prototyping as it requires only a single gradient calculation. To replace the LRA with a function whose gradients are better behaved, we first decompose the perturbed adjacency matrix \(\pertadj = \adj + \diffadj\) and, thus, only need gradients for \(\diffadj\). Next, we notice that the eigenvectors of \(\adj\) usually have few large components. Perturbations along those principal dimensions are representable by the eigenvectors, hence most likely are neither filtered out nor impact the eigenvectors. Knowing this, we approximate the LRA in a tractable manner by element-wise multiplication of \(\diffadj\) with weights that quantify how well an edge aligns with the principal dimensions (details in \autoref{sec:appendix:defenses}). In short we replace \(\operatorname{LRA}(\adj + \diffadj)\) with \(\operatorname{LRA}(\adj) + \diffadj \circ \operatorname{Weight}(\adj)\), which admits useful gradients. This approach carries over to other attacks such as Nettack -- we can incorporate the weights into its score function to avoid selecting edges that will be filtered out.

\textbf{Example 2 -- ProGNN.}
While we approached SVD-GCN with a theoretical insight, breaking a composite defense like ProGNN requires engineering and tinkering. When attacking ProGNN with PGD and transferring the perturbations to poisoning we observe that the perturbations are only effective if the model is trained with the same random seed. This over-sensitivity can be avoided by employing lesson (5) in \autoref{sec:methodology}. As ProGNN is very expensive to train due to its nuclear norm regularizer, we drop that term when training the set of auxiliary models without hurting attack strength. For unrolling the training we again drop the nuclear norm regularizer since it is non-differentiable. Sometimes PGD does not find a state with high attack loss, which can be alleviated by random restarts. As Meta-PGD optimization quickly stalls, we initialize it with a strong perturbation found by Meta-PGD on GCN. All of these tricks combined are necessary to successfully attack ProGNN.

\textbf{Effort.}
Breaking Jaccard-GCN (and SVD-GCN) required around half an hour (resp. three days) of work for the initial proof of concept. Some other defenses require various adjustments that need to be developed over time, but reusing those can quickly break even challenging defenses. It is difficult to quantify this effort, but it can be greatly accelerated by adopting our lessons learned in \autoref{sec:methodology}. In any case, we argue that authors proposing a new defense must put in reasonable effort to break it.

\section{Evaluation of adaptive attacks}\label{sec:adaptive_attacks}

First, we provide details on the experimental setup and used metrics. We then report the main results and findings. We refer to \autoref{sec:appendix:attack_overview} for details on the base attacks, including our Greedy Brute Force and Meta-PGD approaches. We provide the code, configurations, and a collection of perturbed graphs on the project website linked on the first page.

\textbf{Setup.}
We use the two most widely used datasets in the literature, namely Cora~ML~\citep{bojchevski2018deep} and Citeseer~\citep{giles1998citeseer} (details in \autoref{sec:appendix:adaptive_attacks}). Unfortunately, larger datasets are barely possible since most defenses are not very scalable. Still, in \autoref{sec:appendix:scalability}, we discuss scalability and apply an adaptive attack to arXiv (170k nodes)~\citep{hu_open_2020}. We repeat the experiments for five different data splits (10\% training, 10\% validation, 80\% testing) and report the means and variances. We use an internal cluster with Nvidia GTX 1080Ti GPUs. Most experiments can be reproduced within a few hours. However, the experiments with ProGNN and GRAND will likely require several GPU days.

\textbf{Defense hyperparameters.}
When first attacking the defenses, we observed that many exhibit poor robustness using the hyperparameters provided by their authors. To not accidentally dismiss a defense as non-robust, we tune the hyperparameters such that the clean accuracy remains constant but the robustness w.r.t. adaptive attacks is improved. Still, we run all experiments on the untuned defenses as well to confirm we achieve this goal. In the same way, we also tune the GCN model, which we use as a reference to asses whether a defense has merit. We report the configurations and verify the success of our tuning in \autoref{sec:appendix:models_tuned_vs_original}.

\textbf{Attacks and budget.}
In the \emph{global} setting, we run the experiments for budgets \(\Delta\) of up to 15\% of the total number of edges in the dataset. Due to our (R)AUC metric (see below), we effectively focus on only the lower range of evaluated budgets. We apply FGA and PGD~\citep{xu_topology_2019} for evasion. For poisoning, we transfer the found perturbations and also run Metattack~\citep{zugner_adversarial_2019} and our Meta-PGD. Recall that where necessary, we adapt the attacks to the defenses as outlined in \autoref{sec:methodology} and detailed in \autoref{sec:appendix:defenses}.

In the \emph{local} setting, we first draw sets of 20 target nodes per split with degrees 1, 2, 3, 5, 8-10, and 15-25 respectively (total of 120 nodes). This enables us to study how the attacks affect different types of nodes -- lower degree nodes are often conjectured to be less robust (see also \autoref{sec:appendix:node_degreee}). We then run the experiments for relative budgets \(\Delta\) of up to 200\% of the target node's degree. For example, if a node has 10 neighbors, and the budget \(\Delta=70\%\) then the attacker can change up to \(10 \cdot 0.7=7\) edges. This commonly used setup ensures that we treat both low and high-degree nodes fairly. We use Nettack~\citep{zugner_adversarial_2018}, FGA, PGD, and our greedy brute force attack for evasion. For poisoning, we only transfer the found perturbations. Again, we adapt the attacks to the defenses if necessary.

\begin{wrapfigure}[17]{r}{0.4\linewidth}
    \centering
    \vspace{-0.35cm}
    \includegraphics[width=\linewidth]{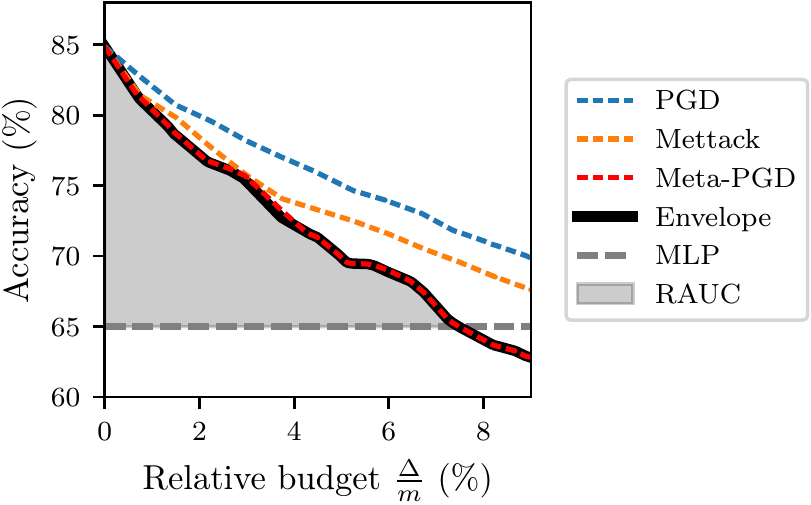}
    \caption{The dotted lines show the test set accuracy per budget after three global poisoning attacks against a tuned GCN on Cora~ML. Taking the envelope gives the solid black robustness curve. The dashed gray line denotes the accuracy of an MLP. The shaded area is the RAUC.\looseness=-1}
    \label{fig:rauc}
\end{wrapfigure}

In alignment with our threat model, we evaluate each found perturbation by the test set accuracy it achieves (\emph{global}) or the ratio of target nodes that remain correctly classified (\emph{local}). For each budget, we choose the strongest attack among all attempts (e.g., PGD, Metattack, Meta-PGD). This gives rise to an envelope curve as seen in \autoref{fig:rauc}. We also include lower budgets as attempts, i.e., we enforce the envelope curve to be monotonically decreasing.

We introduce a rich set of attack characteristics by also transferring the perturbations supporting the envelope curve to every other defense. These transfer attacks then also contribute to the final envelope curve of each defense, but in most cases their contribution is marginal.

\begin{figure}[t]
    \centering
    \captionsetup[subfigure]{justification=centering}
    {
        \captionsetup[subfigure]{margin={1.4cm,0cm}}
        \subcaptionbox{Global, Poisoning}{\includegraphics[scale=0.64]{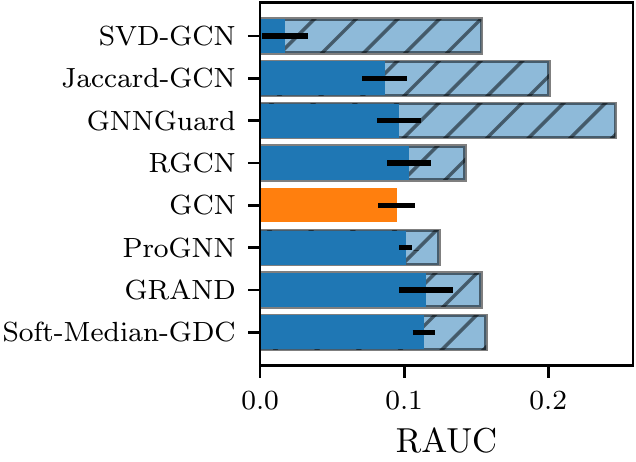}}
    }%
    \hfill
    \subcaptionbox{Global, Evasion}{\includegraphics[scale=0.64]{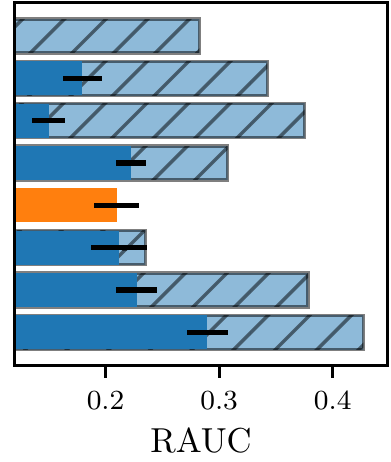}}
    \hfill
    \subcaptionbox{Local, Poisoning}{\includegraphics[scale=0.64]{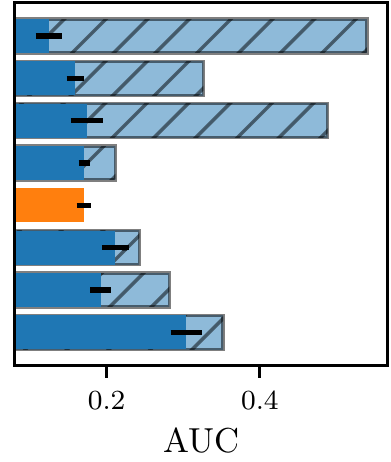}}
    \hfill
    {
        \captionsetup[subfigure]{margin={0cm,1.6cm}}
        \subcaptionbox{Local, Evasion}{\includegraphics[scale=0.64]{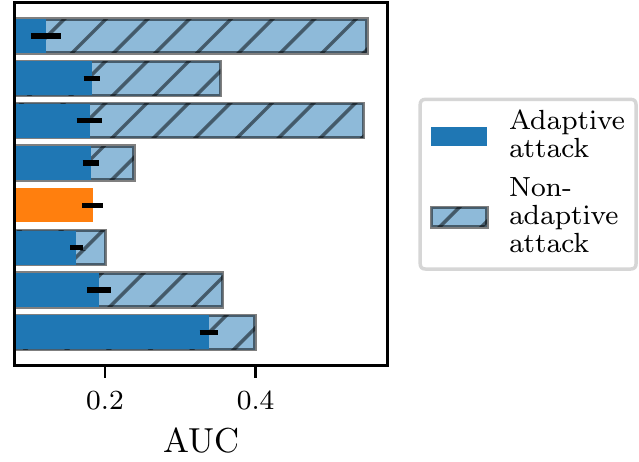}}
    }%
    \caption{Adaptive vs. non-adaptive attacks with budget-agnostic (R)AUC on Cora~ML (c.f. \autoref{fig:superstar}). SVD-GCN (b) is disastrously broken -- our adaptive attacks reach <0.02 (not visible). \autoref{sec:appendix:adaptive_attacks} for Citeseer.}
    \label{fig:improvement_adaptive_auc}
\end{figure}

\textbf{Non-adaptive attacks.} We call any attack ``non-adaptive'' that is not aware of any changes made to the model (including defense mechanisms). Where we report results for a non-adaptive attack (e.g., \autoref{fig:superstar} or \autoref{fig:improvement_adaptive_auc}), we specifically refer to an attack performed on a (potentially linearlized) GCN with commonly used hyperparameters (i.e., untuned). We then apply the perturbed adjacency matrix to the actual defense. In other words, we transfer the adversarial perturbation from a GCN. For our \emph{local} non-adaptive attack, we always use Nettack. In contrast, for our \emph{global} non-adaptive attack, we apply all attacks listed above, and then transfer for each budget the attack which is strongest against the GCN. Due to this ensemble of attacks, our global non-adaptive attack is expected to be slightly stronger than the non-adaptive attacks in most other works.

\begin{figure}[t]
    \centering
    {
        \captionsetup[subfigure]{margin={0.8cm,0cm}}
        \subcaptionbox{Cora~ML, Pois.}{\includegraphics[scale=0.64]{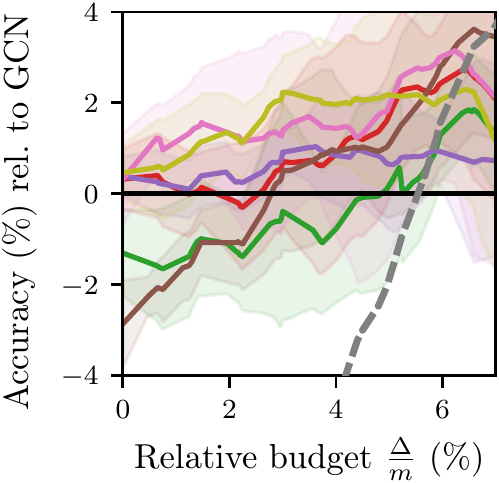}}
    }%
    \hspace{0.04cm}
    \hfill
    \subcaptionbox{Cora~ML, Evas.}{\includegraphics[scale=0.64]{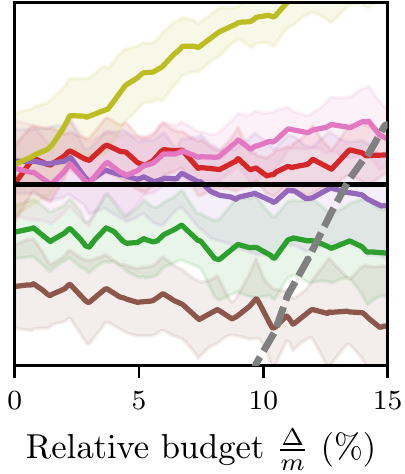}}
    \hfill
    \subcaptionbox{Citeseer, Pois.}{\includegraphics[scale=0.64]{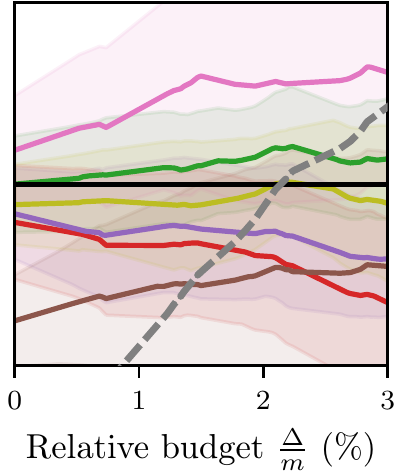}}
    \hfill
    {
        \captionsetup[subfigure]{margin={-2.4cm,0cm}}
        \subcaptionbox{Citeseer, Evas.}{\includegraphics[scale=0.64]{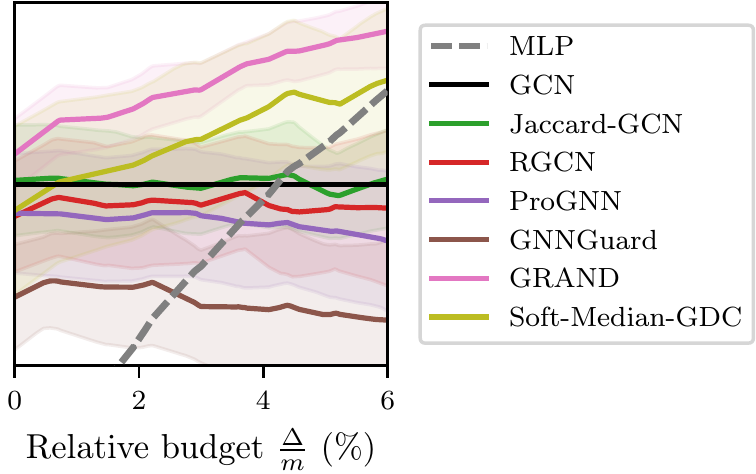}}
    }%
    \caption{Difference (defense -- undefended GCN) of adversarial accuracy for the strongest global attack per budget. Almost half of the defenses perform worse than the GCN. We exclude SVD-GCN since it is catastrophically broken and plotting it would make the other defenses illegible (accuracy <24\% already for a budget of 2\% on Cora~ML). Absolute numbers in \autoref{sec:appendix:adaptive_attacks}.}
    \label{fig:global_mr_envelope}
\end{figure}

\textbf{Area Under the Curve (AUC).}
An envelope curve gives us a detailed breakdown of the empirical robustness of a defense for different adversarial budgets. However, it is difficult to compare different attacks and defenses by only visually comparing their curves in a figure (e.g., see \autoref{fig:global_mr_envelope}). Therefore, in addition to this breakdown per budget, we summarize robustness using the Area Under the Curve (AUC), which is independent of a specific choice of budget \(\Delta\) and also punishes defenses that achieve robustness by trading in too much clean accuracy. Intuitively higher AUCs indicate more robust models, and conversely, lower AUCs indicate stronger attacks.

As our \emph{local} attacks break virtually all target nodes within our conservative maximum budget (see \autoref{sec:appendix:adaptive_attacks}), taking the AUC over all budgets conveniently measures how quick this occurs. However, for \emph{global} attacks, the test set accuracy continues to decrease for unreasonably large budget, and it is unclear when to stop. To avoid having to choose a maximum budget, we wish to stop when discarding the entire tainted graph becomes the better defense. This is fulfilled by the area between the envelope curve and the line signifying the accuracy of an MLP -- a model that is oblivious to the graph structure, at the expense of a substantially lower clean accuracy than a GNN. We call this metric Relative AUC (RAUC) and illustrate it in \autoref{fig:rauc}. More formally, \(\operatorname{RAUC}(c) = \int_0^{b_0} (c(b) - a_\text{MLP}) \mathrm{d} b\)\: s.t. \:\(b \lessgtr b_0 \Longrightarrow c(b) \gtrless a_\text{MLP}\)\: where \(c(\cdot)\) is a piecewise linear robustness per budget curve, and \(a_\text{MLP}\) is the accuracy of the MLP baseline. We normalize the RAUC s.t. 0\% is the performance of an MLP and 100\% is the optimal score (i.e., 100\% accuracy).

\begin{figure}[b]
    \centering
    {
        \captionsetup[subfigure]{margin={0.8cm,0cm}}
        \subcaptionbox{Cora~ML, Pois.}{\includegraphics[scale=0.64]{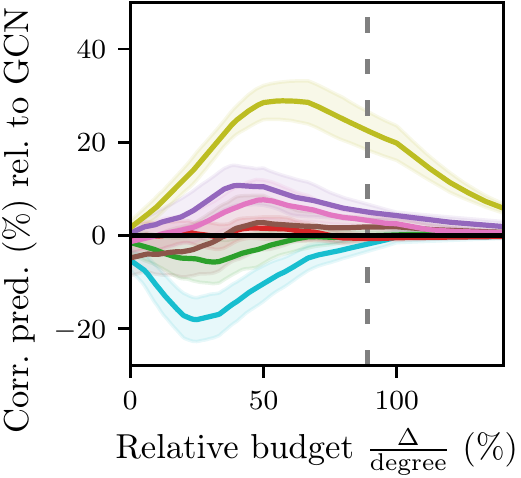}}
    }%
    \hfill
    \subcaptionbox{Cora~ML, Evas.}{\includegraphics[scale=0.64]{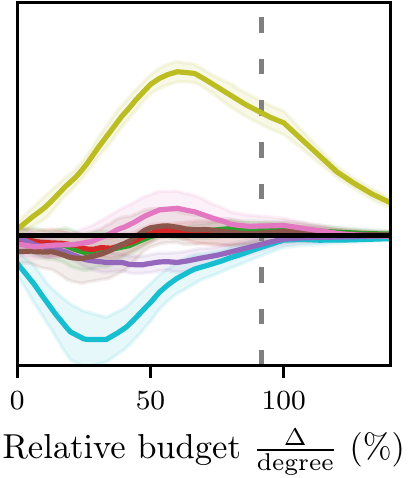}}
    \hfill
    \subcaptionbox{Citeseer, Pois.}{\includegraphics[scale=0.64]{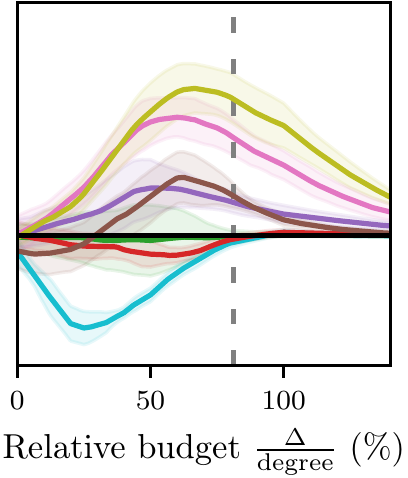}}
    \hfill
    {
        \captionsetup[subfigure]{margin={-2.4cm,0cm}}
        \subcaptionbox{Citeseer, Evas.}{\includegraphics[scale=0.64]{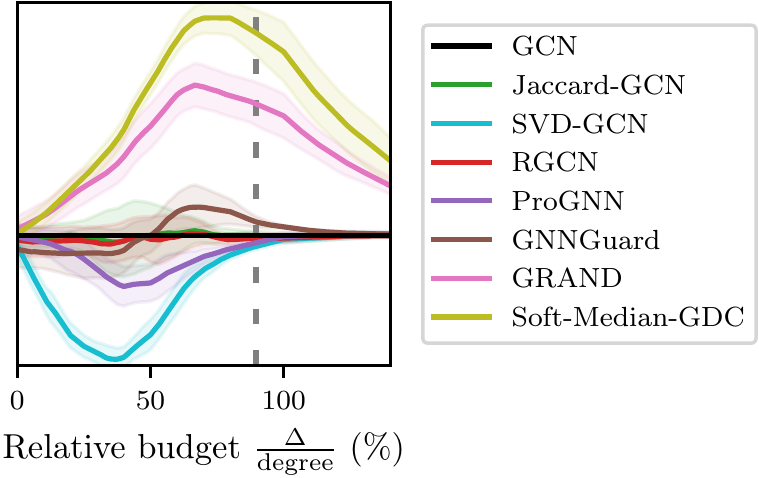}}
    }%
    \caption{Difference (defense -- undefended GCN) of fraction of correct predictions for the strongest local attack per budget. Most defenses show no or only marginal gain in robustness. The dashed vertical line shows where 95\% of nodes for a GCN are misclassified on average. Abs. numbers in \autoref{sec:appendix:adaptive_attacks}.}
    \label{fig:local_mr_envelope}
\end{figure}

\textbf{Finding 1 -- Our adaptive attacks lower robustness by 40\% on average.} 
In \autoref{fig:improvement_adaptive_auc} we compare non-adaptive attacks, the current standard to evaluate defenses, with our adaptive attacks which we propose as a new standard. The achieved (R)AUC in each case drops on average by 40\% (similarly for Citeseer, see \autoref{sec:appendix:adaptive_attacks}). In other words, the reported robustness in the original works proposing a defense is roughly 40\% too optimistic. We confirm a statistically significant drop (\(p < 0.05\)) with a one-sided t-test in 85\% of all cases. Considering adversarial accuracy for (small) fixed adversarial budget (\autoref{fig:superstar}) instead of the summary (R)AUC over all budgets tells the same story: non-adaptive attacks are too weak to be reliable indicators of robustness and adaptive attacks massively shrink the alleged robustness gains.

\textbf{Finding 2 -- Structural robustness of GCN is not easily improved.} 
In \autoref{fig:global_mr_envelope} (global) and \autoref{fig:local_mr_envelope} (local) we provide a more detailed view for different adversarial budgets and different graphs. For easier comparison we show the accuracy relative to the undefended GCN baseline. Overall, the decline is substantial. Almost half of the examined defenses perform worse than GCN and most remaining defenses neither meaningfully improve nor lower the robustness (see also \autoref{fig:superstar} and \autoref{fig:rauc}). GRAND and Soft-Medoid-GCN retain robustness in some settings, but the gains are smaller than reported.\looseness=-1

\textbf{Finding 3 -- Defense effectiveness depends on dataset.}
As we can see in \autoref{fig:global_mr_envelope} and \autoref{fig:local_mr_envelope}, our ability to circumvent specific defenses tends to depend on the dataset. It appears that some defenses are more suited for different datasets. For example, GRAND seems to be a good choice for Citeseer while it is not as strong on Cora~ML. The results for local attacks (\autoref{fig:local_mr_envelope}) paint a similar picture, here we see that Cora~ML is more difficult to defend. This points to another potentially problematic pitfall: most defenses are developed only using these two datasets as benchmarks. Is robustness even worse on other graphs? We leave this question for future work.

\begin{wrapfigure}[13]{r}{0.4\textwidth}
    \centering
    \vspace{-0.61cm}
    \includegraphics[width=\linewidth]{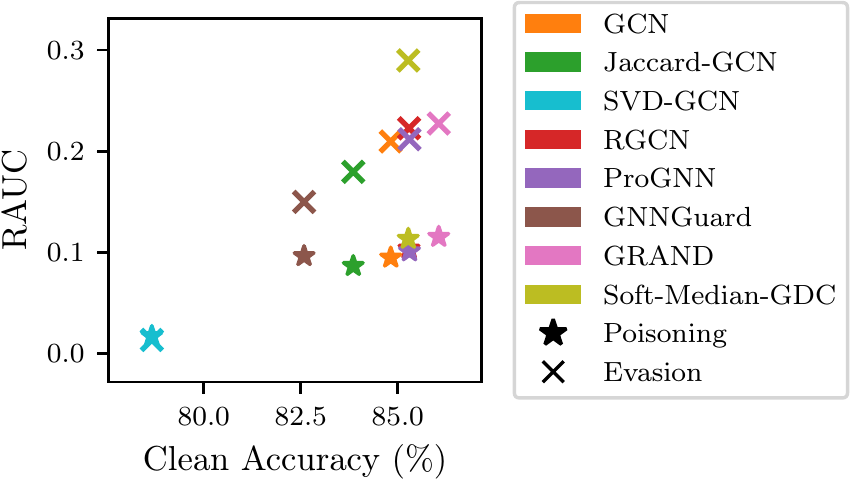}
    \caption{Model accuracy vs. RAUC of the strongest global attacks on Cora~ML. We do not observe a robustness accuracy trade-off, but even find models with higher accuracy to be more robust.}
    \label{fig:accuracy_robustness_tradeoff}
\end{wrapfigure}

\textbf{Finding 4 -- No trade-off between accuracy and robustness for structure perturbations.}
Instead, \autoref{fig:accuracy_robustness_tradeoff} shows that defenses with high clean accuracy also exhibit high RAUC, i.e., are more robust against our attacks. This appears to be in contrast to the image domain \citep{tsipras_robustness_2019}. However, we cannot exclude that future more powerful defenses might manifest this trade-off in the graph domain.

\begin{figure}
    \subcaptionbox{Node degree}{\includegraphics[scale=0.7]{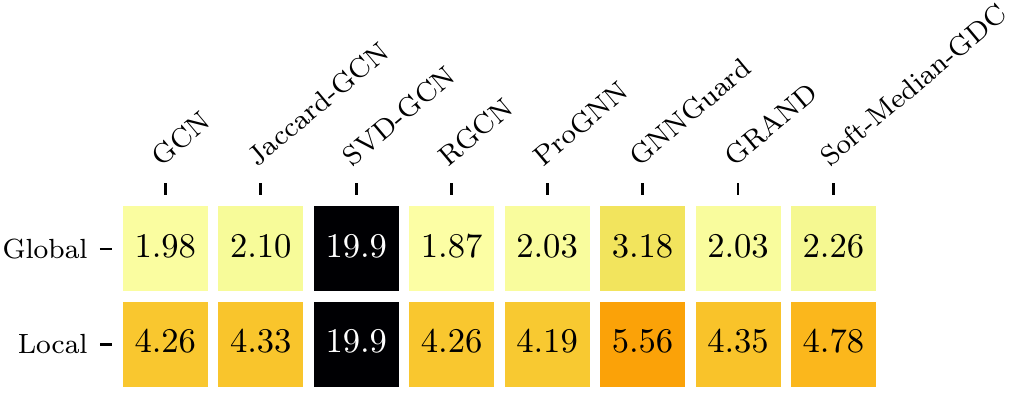}}
    {
        \captionsetup[subfigure]{margin={0cm,0.7cm}}
        \subcaptionbox{Ratio of removed edges}{\includegraphics[scale=0.7]{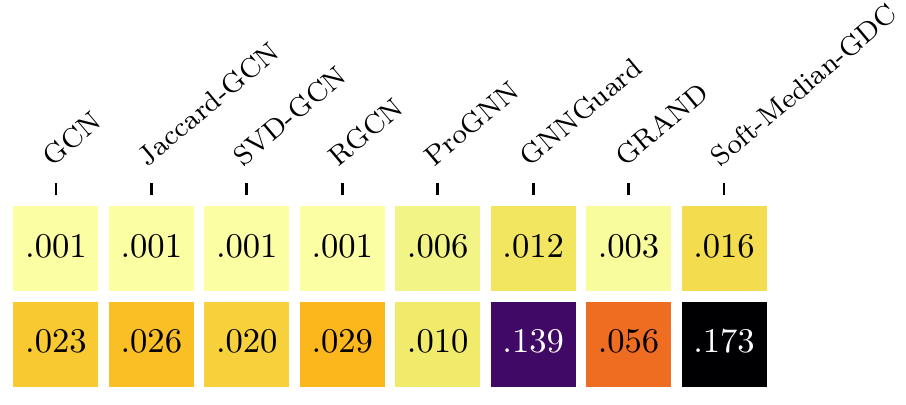}}
    }
    \caption{Exemplary metrics characterizing the attack vector our strongest attacks, which are those visible in \autoref{fig:appendix:global_envelope_support} and \autoref{fig:appendix:local_envelope_support}. We give a more elaborate study of attack characteristics in \autoref{sec:appendix:attack_characteristics}.}
    \label{fig:attack_char_matrix}
\end{figure}

\textbf{Finding 5 -- Necessity of adaptive attacks.} In \autoref{fig:attack_char_matrix}, we show two exemplary characteristics of how an adaptive attack bypasses defensive measures. First, to attack SVD-GCN, it seems particularly effective to insert connections to high-degree nodes. Second, for GNNGuard, GRAND and Soft-Median-GDC it is disproportionally helpful to delete edges. These examples illustrate why the existence of a one-fits-all perturbation which circumvents all possible defenses is unlikely. Instead, an adaptive attack is necessary to properly assess a defense's efficacy since different models are particularly susceptible to different perturbations.

\textbf{Additional analysis.}
During this project, we generated a treasure trove of data. We perform a more in-depth analysis of our attacks in the appendix. First, we study how node degree affects attacks (see \autoref{sec:appendix:node_degreee}). For local attacks, the required budget to misclassify a node is usually proportional to the node's degree. Global attacks tend to be oblivious to degree and uniformly break nodes. Next, we perform a breakdown of each defense in terms of the sensitivity to different attacks (see \autoref{sec:appendix:attack_success}). In short, global attacks are dominated by PGD for evasion and Metattack/Meta-PGD for poisoning with the PM or TLM loss. For local, our greedy brute-force is most effective, rarely beaten by PGD and Nettack. Finally, we analyze the properties of the adversarial edges in terms of various graph statistics such as edge centrality and frequency spectra (see \autoref{sec:appendix:attack_characteristics}\, \autoref{sec:appendix:spectral_attack_characteristics}).

\section{Robustness unit test}\label{sec:black_box_attack}

\begin{figure}[b]
    \centering
    {
        \captionsetup[subfigure]{margin={1.7cm,0cm}}
        \subcaptionbox{Poisoning}{\includegraphics[scale=0.75]{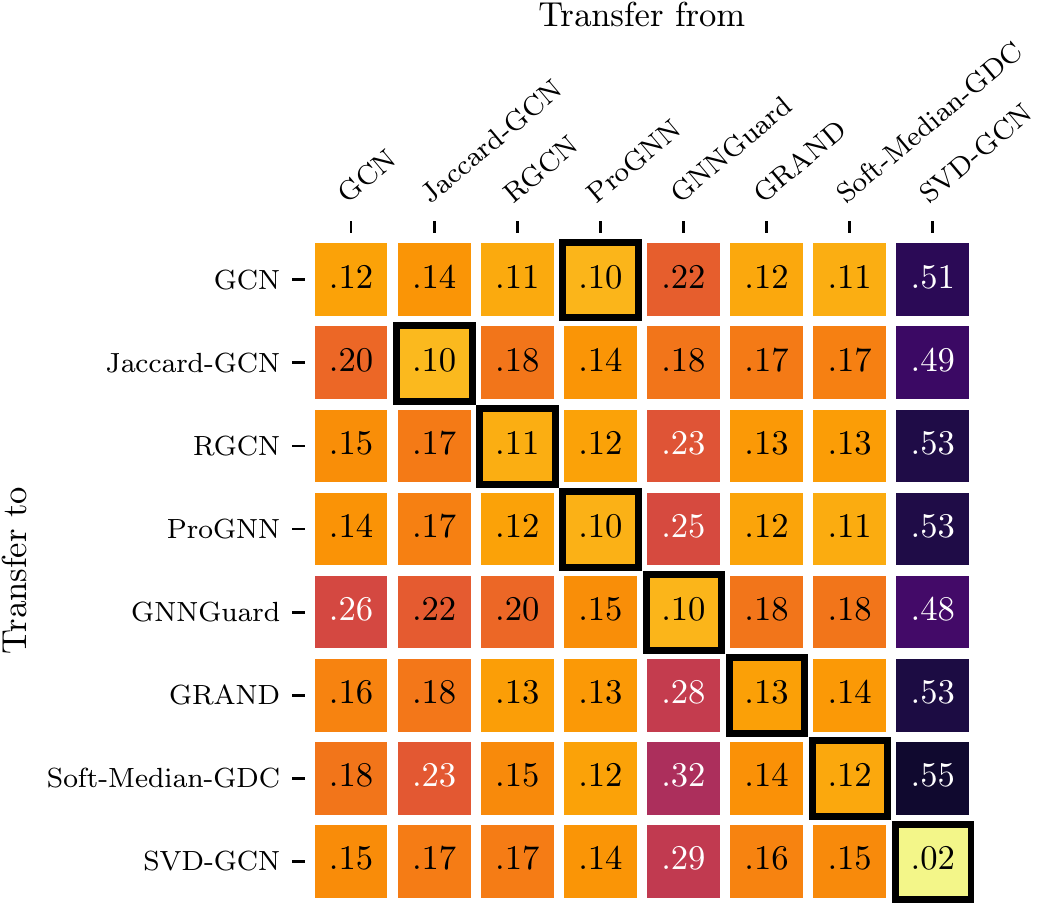}}
    }%
    \hfill
    {
        \captionsetup[subfigure]{margin={-0.5cm,0cm}}
        \subcaptionbox{Evasion}{\includegraphics[scale=0.75]{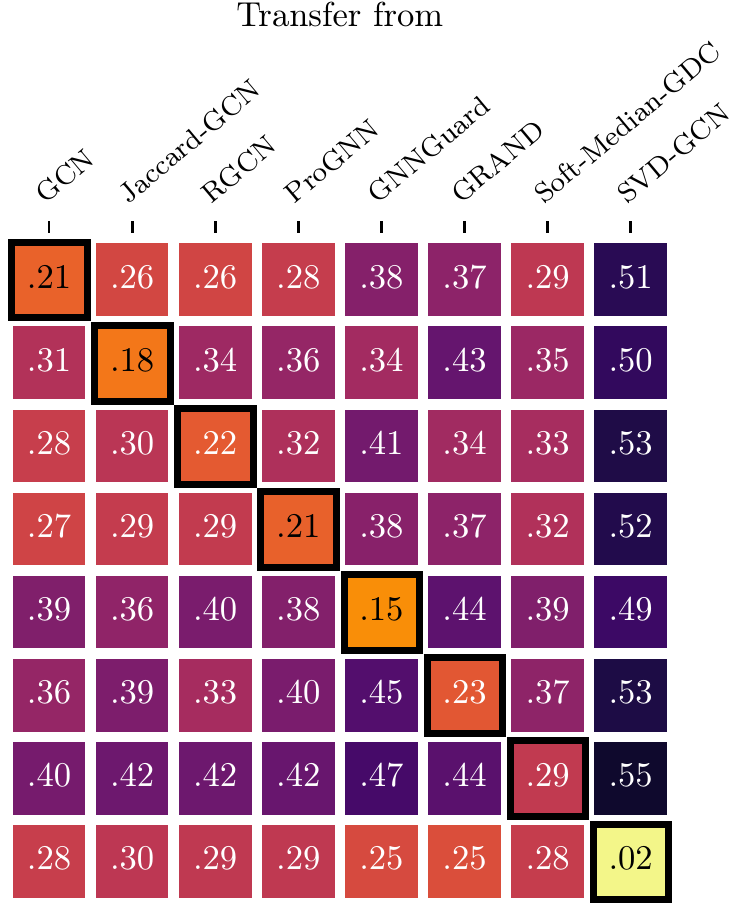}}
    }%
    \caption{RAUC for the transfer of the strongest global adaptive attacks on Cora~ML between models. The columns contain the models for which the adaptive attacks were created. The rows contain the RAUC after the transfer. With only one exception, adaptive attacks (diagonal) are most effective.}
    \label{fig:transfer}
\end{figure}

Next we systematically study how well the attacks transfer between defenses, as introduced in the \emph{attacks and budget} paragraph in \autoref{sec:adaptive_attacks}. In \autoref{fig:transfer}, we see that in 15 out of 16 cases the adaptive attack is the most effective strategy (see main diagonal). However for many defenses, there is often a source model or ensemble of source models (for the latter see \autoref{sec:appendix:ensemble_transfer}) which forms a strong transfer attack.

Motivated by the effectiveness of transfer attacks (especially if transferring from ProGNN~\citep{jin_graph_2020}), we suggest this set of perturbed graphs to be used as a bare minimum robustness unit test: one can probe a new defense by testing against these perturbed graphs, and if there exists at least one that diminishes the robustness gains, we can immediately conclude that the defense is not robust in the worst-case -- without the potentially elaborate process of designing a new adaptive attack. We provide instructions on how to use this collection in the accompanying code.

Nevertheless, we cannot stress enough that this collection does not replace a properly developed adaptive attack. For example, if one would come up with SVD-GCN and would use our collection (excluding the perturbed graphs for SVD-GCN) the unit test would partially pass. However, as we can see in e.g., \autoref{fig:improvement_adaptive_auc}, SVD-GCN can be broken with an -- admittedly very distinct -- adaptive attack.

\section{Related work}\label{sec:related_work}

Excluding attacks on undefended GNNs, previous works studying adaptive attacks in the graph domain are scarce. The recently proposed graph robustness benchmark~\citep{zheng_graph_2021} also only studies transfer attacks. Such transfer attacks are so common in the graph domain that their usage is often not even explicitly stated, and we find that the perturbations are most commonly transferred from Nettack or Metattack (both use a linearized GCN). Other times, the authors of a defense only state that they use PGD~\citep{xu_topology_2019} (aka ``topology attack'') without further explanations. In this case, the authors most certainly refer to a PGD transfer attack on a GCN proxy. They almost never apply PGD to their actual defense, which would yield an adaptive attack (but possibly weak, see \autoref{sec:methodology} for guidance).

An exception where the defense authors study an adaptive attack is SVD-GCN~\citep{entezari_all_2020}. Their attack collects the edges flipped by Nettack in a difference matrix \(\diffadj\), replaces its most significant singular values and vectors with those from the clean adajcency matrix \(\adj\), and finally adds it to \(\adj\). Notably, this yields a dense continuous perturbed adjacency matrix. While their SVD-GCN is susceptible to these perturbations, the results however do not appear as catastrophic as with our adaptive attacks, despite their severe violation of our threat model (see \autoref{sec:preliminaries}). \citet{geisler_robustness_2021} are another exception where gradient-based greedy and PGD attacks are directly applied to their Soft-Median-GDC defense, making them adaptive. Still, our attacks manage to further reduce their robustness estimate.

\section{Discussion}\label{sec:discussion}

We hope that the adversarial learning community for GNNs will reflect on the bitter lesson that evaluating adversarial robustness is not trivial. We show that on average adversarial robustness estimates are overstated by 40\%. To ease the transition into a more reliable regime of robustness evaluation for GNNs we share our recipe for successfully designing strong adaptive attacks.

Using adaptive (white-box) attacks is also interesting from a security perspective. If a model successfully defends such strong attacks, it is less likely to have remaining attack vectors for a real-world adversary. Practitioners can use our methodology to evaluate their models in hope to avoid an arms race with attackers. Moreover, the white-box assumption lowers the chance that real-world adversaries can leverage our findings, as it is unlikely that they have perfect knowledge.

We also urge for caution since the attacks only provide an upper bound (which with our attacks is now 40\% tighter). Nevertheless, we argue that the burden of proof that a defense is truly effective should lie with the authors proposing it. Following our methodology, the effort to design a strong adaptive attack is reduced, so we advocate for adaptive attacks as the gold-standard for future defenses.

\begin{ack}
This research was supported by the Helmholtz Association under the joint research school “Munich School for Data Science -- MUDS“.
\end{ack}

\clearpage
\bibliography{references}

\section*{Checklist}

\begin{enumerate}

\item For all authors...
\begin{enumerate}
  \item Do the main claims made in the abstract and introduction accurately reflect the paper's contributions and scope?
    \answerYes{}
  \item Did you describe the limitations of your work?
    \answerYes{See \autoref{sec:discussion}.}
  \item Did you discuss any potential negative societal impacts of your work?
    \answerYes{See \autoref{sec:discussion}.}
  \item Have you read the ethics review guidelines and ensured that your paper conforms to them?
    \answerYes{}
\end{enumerate}

\item If you are including theoretical results...
\begin{enumerate}
  \item Did you state the full set of assumptions of all theoretical results?
    \answerNA{}
  \item Did you include complete proofs of all theoretical results?
    \answerNA{}
\end{enumerate}

\item If you ran experiments...
\begin{enumerate}
  \item Did you include the code, data, and instructions needed to reproduce the main experimental results (either in the supplemental material or as a URL)?
    \answerYes{See \autoref{sec:adaptive_attacks}.}
  \item Did you specify all the training details (e.g., data splits, hyperparameters, how they were chosen)?
    \answerYes{See \autoref{sec:adaptive_attacks}, \autoref{sec:appendix:models_tuned_vs_original} and provided code.}
  \item Did you report error bars (e.g., with respect to the random seed after running experiments multiple times)?
    \answerYes{All experiments are repeated for five random data splits.}
  \item Did you include the total amount of compute and the type of resources used (e.g., type of GPUs, internal cluster, or cloud provider)?
    \answerYes{See beginning of \autoref{sec:adaptive_attacks}.}
\end{enumerate}

\item If you are using existing assets (e.g., code, data, models) or curating/releasing new assets...
\begin{enumerate}
  \item If your work uses existing assets, did you cite the creators?
    \answerYes{}
  \item Did you mention the license of the assets?
    \answerNo{}
  \item Did you include any new assets either in the supplemental material or as a URL?
    \answerYes{See beginning of \autoref{sec:adaptive_attacks}.}
  \item Did you discuss whether and how consent was obtained from people whose data you're using/curating?
    \answerNA{}
  \item Did you discuss whether the data you are using/curating contains personally identifiable information or offensive content?
    \answerNA{}
\end{enumerate}

\item If you used crowdsourcing or conducted research with human subjects...
\begin{enumerate}
  \item Did you include the full text of instructions given to participants and screenshots, if applicable?
    \answerNA{}
  \item Did you describe any potential participant risks, with links to Institutional Review Board (IRB) approvals, if applicable?
    \answerNA{}
  \item Did you include the estimated hourly wage paid to participants and the total amount spent on participant compensation?
    \answerNA{}
\end{enumerate}

\end{enumerate}

\appendix

\counterwithin{figure}{section}
\counterwithin{table}{section}
\counterwithin{equation}{section}
\counterwithin{algorithm}{section}

\section{Attacks overview}\label{sec:appendix:attack_overview}

In this section, we make the ensemble of attacks explicit and explain essential details. We then adapt these attack primitives to circumvent the defense mechanisms (see \autoref{sec:appendix:defenses}).

\textbf{Global evasion attacks.}
The goal of a global attack is to provoke the misclassification of a large fraction of nodes (i.e., the test set) jointly, crafting a single perturbed adjacency matrix. For evasion, we use \emph{(1) the Fast Gradient Attack (FGA)} and \emph{(2) Projected Gradient Descent (PGD)}. In FGA, we calculate the gradient towards the entries of the clean adjacency matrix \(\nabla_{\adj} \loss_{\text{attack}} (\model(\adj, \feat), \lab)\) and then flip the highest-ranked edges at once s.t. we exhaust the budget \(\Delta\). In contrast, PGD requires multiple gradient updates since it uses gradient ascent (see \autoref{sec:preliminaries} or explanation below for Meta-PGD). We deviate from the PGD implementation of \citet{xu_topology_2019} is two ways: (I) we adapt the initialization of the perturbation before the first attack gradient descent step and (II) we adjust the final sampling of \(\pertadj\). See below for more details.

\textbf{Global poisoning attacks.}
We either (a) transfer the perturbation \(\pertadj\) found by evasion attack (1) or (2) and use it to poison training, or (b) differentiate through the training procedure by unrolling it, thereby obtaining a meta gradient. The latter approach is taken by both \emph{(3) Metattack}~\citep{zugner_adversarial_2019} and \emph{(4) our Meta-PGD}. Metattack greedily flips a single edge in each iteration and then obtains a new meta gradient at the changed adjacency matrix. In Meta-PGD, we follow the same relaxation as \citet{xu_topology_2019} (see below as well as \autoref{sec:preliminaries}) and obtain meta gradients at the relaxed adjacency matrices. In contrast to the greedy approach of Metattack, Meta-PGD is able to revise early decisions later on.

\textbf{Meta-PGD.}
Next, we explain the details of Meta-PGD and we present the pseudo code for reference in \autoref{algo:appendix:meta_pgd}. Recall that the discrete edges are relaxed \(\{0,1\} \to [0, 1]\) and that the ``weight'' of the perturbation reflects the probability of flipping the respective edge. 

\begin{algorithm}[h]
    \small 
    \caption{Meta-PGD}
    \label{algo:appendix:meta_pgd}
    \begin{algorithmic}[1]
        \STATE {\bfseries Input:} Adjacency matrix \(\adj\), node features \(\feat\), labels \(\lab\), GNN \(f_{\theta}(\cdot)\), loss \(\loss_{\text{attack}}\)
        \STATE {\bfseries Parameters:} Budget\ \(\Delta\), iterations \(E\), learning rates \(\alpha_t\)
        \STATE Initialize \(\mathbf{P}_0 \in \mathbb{R}^{n \times n}\)
        \FOR{\(t \in \{1,2, \dots, E\}\)}
        \STATE Step \(\mathbf{P}^{(t)} \leftarrow \mathbf{P}^{(t-1)} + \alpha_t \nabla_{\mathbf{P}^{(t-1)}} \left[ \loss_{\text{attack}} \left( f \big(\adj + \mathbf{P}^{(t-1)}, \feat; \, \theta = \operatorname{train}(\adj + \mathbf{P}^{(t-1)}, \feat, \lab) \big), \lab \right) \right]\)
        \STATE Projection \(\mathbf{P}^{(t)} \leftarrow \Pi_{\|\mathbb{E}[\adj + \mathbf{P}^{(t)}] - \adj\|_0 \le 2\Delta} (\mathbf{P}^{(t)})\)
        \ENDFOR
        \STATE Sample \(\pertadj\) s.t.\ \(\|\pertadj - \adj\|_0 \le 2\Delta\)
        \STATE Return \(\pertadj\)
    \end{algorithmic}
\end{algorithm}

In the first step of Meta-PGD, we initialize the perturbation (line 3). In contrast to \citet{xu_topology_2019}'s suggestion, we find that initializing the perturbation with the zero matrix can cause convergence issues. Hence, we alternatively initialize the perturbation with \(\pertadj\) from an attack on a different model (see also lesson learned \#8 in \autoref{sec:methodology}).

In each attack iteration, a gradient ascent step is performed on the relaxed perturbed adjacency matrix \(\pertadj^{(t-1)} = \adj + \mathbf{P}^{(t-1)}\) (line 5). For obtaining the meta gradient through the training process, the training is unrolled. For example, with vanilla gradient descent for training \(f_{\theta}(\adj, \feat) = f(\adj, \feat; \theta)\), the meta gradient resolves to
\begin{equation}
    \resizebox{0.9\linewidth}{!}{\(
        \nabla_{\mathbf{P}^{(t-1)}} \left( \loss_{\text{attack}} \left[ f\Big(\adj + \mathbf{P}^{(t-1)}, \feat; \theta = \theta_0 - \eta \sum\limits_{k=1}^{E_{\text{train}}} \nabla_{\theta_{k-1}} \loss_{\text{train}} [ f(\adj + \mathbf{P}^{(t-1)}, \feat; \theta=\theta_{k-1}), \lab ] \Big), \lab \right] \right)
    \)}
\end{equation}
with number of training epochs \(E_{\text{train}}\), fixed training learning rate \(\eta\), and parameters after (random) initialization \(\theta_{0}\). Notice that to obtain our variant of non-meta PGD, it suffices to replace the gradient computation in line 5 with \(\nabla_{\mathbf{P}^{(t-1)}} \left[ \loss_{\text{attack}}( \model(\adj + \mathbf{P}^{(t-1)}, \feat), \lab) \right]\).

Thereafter in line 6, the perturbation is projected such that in expectation the budget is obeyed, i.e., \(\Pi_{\|\mathbb{E}[\adj + \mathbf{P}^{(t)}] - \adj\|_0 \le 2\Delta}\). First, the projection clips \(\adj + \mathbf{P}^{(t-1)}\) to be in \([0,1]\). If the budget is violated after clipping, it solves
\begin{equation}
    \arg\min_{\hat{\mathbf{P}}^{(t)}} \|\hat{\mathbf{P}}^{(t)} - \mathbf{P}^{(t)}\|_2 \qquad \text{ s.t. } \quad \adj + \hat{\mathbf{P}}^{(t)} \in [0,1]^{n \times n} \text{  and  } \sum |\hat{\mathbf{P}}^{(t)}| \le 2\Delta
\end{equation}

After the last iteration (line 7), each element of \(\mathbf{P}^{(t)}\) is interpreted as a probability and multiple perturbations are sampled accordingly. The strongest drawn perturbed adjacency matrix (in terms of attack loss) is chosen as \(\pertadj\). Specifically, in contrast to \citep{xu_topology_2019}, we sample \(K=100\) potential solutions that all obey the budget \(\Delta\) and then choose the one that maximizes the attack loss \(\loss_{\text{attack}}\).

\textbf{Local attacks.}
For local attacks we only run evasion attacks, and then transfer them to poisoning. This is common practice (e.g., see \citet{zugner_adversarial_2018} or \citet{li2021adversarial}). The attacks we use are \emph{(1) FGA}, \emph{(2) PGD}, \emph{(3) Nettack~\citep{zugner_adversarial_2018}}, and a \emph{(4) Greedy Brute Force} attack. Nettack greedily flips the best edges considering a linearized GCN, whose weights are either specially trained or taken from the attacked defense. In contrast, in each iteration, our Greedy Brute Force attack flips the current worst-case edge for the attacked model. It determines the worst-case perturbation by evaluating the model for every single edge flip. Notice that all examined models use two propagation steps, so we only consider all potential edges adjoining the target node or its neighbors\footnote{Due to GCN-like normalization (see \autoref{sec:appendix:defenses}), the three-hop neighbors need to be considered to be exhaustive. However, it is questionable if perturbing a neighbor three hops away is ever the strongest perturbation there is.}. Importantly, Greedy Brute Force is adaptive for any kind of model. Runtime-wise, the algorithm evaluates the attacked model \(\mathcal{O}(\Delta n d)\) times with the number of nodes \(n\) and the degree of the target node \(d\). We provide pseudo code in \autoref{algo:appendix:gbf}.

\begin{algorithm}[h]
    \small 
    \caption{Greedy Brute Force}
    \label{algo:appendix:gbf}
    \begin{algorithmic}[1]
        \STATE {\bfseries Input:} Target node \(i\), adjacency matrix \(\adj\), node features \(\feat\), labels \(\lab\), GNN \(f_{\theta}(\cdot)\), loss \(\loss_{\text{attack}}\)
        \STATE {\bfseries Parameter:} Budget\ \(\Delta\)
        \STATE Initialize \(\pertadj^{(0)} = \adj\)
        \FOR{\(t \in \{1,2, \dots, \Delta\}\)}
        \FOR{potential edge \(e\) adjoining \(i\) or any of \(i\)'s direct neighbors}
        \STATE Flip edge \(\pertadj^{(t)} \leftarrow \pertadj^{(t-1)} \pm e\)
        \STATE Remember best \(\pertadj^{(t)}\) in terms of \(\loss_{\text{attack}}(\model(\pertadj^{(t)}, \feat), \lab)\)
        \IF{node \(i\) is missclassifed}
            \STATE Return \(\pertadj^{(t)}\)
        \ENDIF
        \ENDFOR
        \STATE Recover best \(\pertadj^{(t)}\)
        \ENDFOR
        \STATE Return \(\pertadj_\Delta\)
    \end{algorithmic}
\end{algorithm}

\textbf{Unnoticeability}
typically serves as a proxy to ensure that the label of an instance (here node) has not changed. In the image domain, it is widely accepted that a sufficiently small perturbation of the input image w.r.t. an \(L_p\)-norm is unnoticeable (and similarly for other threat models such as rotation). For graphs the whole subject of unnoticeability is more nuanced. The only constraint we use is the number of edge insertions/deletion, i.e., an \(L_0\)-ball around the clean adjacency matrix. 

The only additional unnoticeability constraint proposed in the literature compares the clean and perturbed graph under a power law assumption on the node degrees~\citep{zugner_adversarial_2018}. However, we do not include such a constraint since (1) the degree distribution is only one (arbitrary) property to distinguish two graphs. (2) The degree distribution is a global property with an opaque relationship to the local class labels in node classification. (3) As demonstrated in \citet{zugner_adversarial_2019}, enforcing an indistinguishable degree distribution only has a negligible influence on attack efficacy, i.e., their gradient-based/adaptive attack conveniently circumvents this measure. Thus, we argue that enforcing such a constraint is similar to an additional (weak) defense measure and is not the focus of this work. Finally, since many defense (and attack) works in the literature considering node-classification (including the ones we study) also only use an \(L_0\)-ball constraint as a proxy for unnoticeability, we do the same for improved consistency. Out of scope are also other domains, like combinatorial optimization, where unnoticeability is not required since the true label of the perturbed instance is known~\citep{geisler_rco2022}.

\newpage
\section{Defense taxonomy}\label{sec:appendix:taxonomy}

Next, we give further details behind our reasoning on how to categorize defenses for GNNs. Our taxonomy extends and largely follows \citet{GNNBook-ch8-gunnemann}'s. The three main categories are \emph{improving the graph} (\autoref{sec:appendix:taxonomy_graph}), \emph{improving the training} (\autoref{sec:appendix:taxonomy_training}), and \emph{improving the architecture} (\autoref{sec:appendix:taxonomy_architecture}). We assign each defense to the category that fits best, even though some defenses additionally include ideas fitting into other categories as well. For the assignment of defenses see \autoref{tab:categorization}.

\subsection{Improving the graph}\label{sec:appendix:taxonomy_graph}

With this category, we refer to all kinds of preprocessing of the graph. Alternatively, some approaches make the graph learnable with the goal of improved robustness. In summary, this category addresses changes that take place \emph{prior} to the GNN (i.e., any message passing). We further distinguish \emph{(1) unsupervised} and \emph{(2) supervised} approaches.

\textbf{Unsupervised.}
Any improvements that are not entangled with a learning objective, i.e., pure preprocessing, usually arising from clues found in the node features and graph structure. For example, Jaccard-GCN~\citep{wu_adversarial_2019} filters out edges based on the Jaccard similarity of node features, while SVD-GCN~\citep{entezari_all_2020} performs a low-rank approximation to filter out high-frequency perturbations. Most other approaches from this category exploit clues from features and structure simultaneously.

\textbf{Supervised.}
These graph improvements are entangled with the learning objective by making the adjacency matrix learnable, often accompanied by additional regularization terms that introduce expert assumptions about robustness. For example, ProGNN~\citep{jin_graph_2020} treats the adjacency matrix like a learnable parameter, and adds loss terms s.t. it remains close to the original adjacency matrix and exhibits properties which are assumed about clean graphs like low-rankness.

\subsection{Improving the training}\label{sec:appendix:taxonomy_training}

These approaches improve training -- without changing the architecture -- s.t. the learned parameters \(\params^*\) of the GNN exhibit improved robustness. In effect, the new training ``nudges'' a regular GNN towards being more robust. We distinguish \emph{(1) robust training} and \emph{(2) further training principles}.

\textbf{Robust training.}
Alternative training schemes and losses which reward the correct classification of synthetic adversarial perturbations of the training data. With this category, \citet{GNNBook-ch8-gunnemann} targets both straightforward adversarial training and losses stemming from certificates (i.e., improving certifiable robustness). Neither approach is interesting to us: the former is discussed in \autoref{sec:appendix:adversarial_training}, and the latter targets provable robustness which does not lend itself to empirical evaluation.

\textbf{Further training principles.}
This category is distinct from robust training due to the lack of a clear mathematical definition of the training objective. It mostly captures augmentations~\citep{feng_graph_2021, jin_power_2021, regol_node_2022, tang_transferring_2020, zheng_robust_2020} or alternative training schemes~\citep{chang_not_2021, elinas_variational_2020, you_graph_2020, zhuang_defending_2022} that encourage robustness. A simple example for such an approach is to pre-train the GNN weights on perturbed graphs~\citep{tang_transferring_2020}. Another recurring theme is to use multiple models during training and then, e.g., enforce consistency among them~\citep{chang_not_2021}.

\subsection{Improving the architecture}\label{sec:appendix:taxonomy_architecture}

Even though there are some exceptions (see sub-category \emph{(2) miscellaneous}), the recurring theme in this category is to somehow weight down the influence of some edges adaptively for each layer or message passing aggregation. We refer to this type of improved architecture with \emph{(1) adaptively weighting edges}. We further distinguish between approaches that are \emph{(a) rule-based}, \emph{(b) probabilistic}, or use \emph{(c) robust aggregation}.

\emph{Rule-based} approaches typically use some metric~\citep{jin_node_2021, zhang_gnnguard_2020}, alternative message passing~\citep{liu_graph_2021, liu_elastic_2021}, or an auxiliary MLP~\citep{zhang_feature-importance-aware_2020} to filter out alleged adversarial edges. \emph{Probabilistic} approaches either work with distributions in the latent space~\citep{zhu_robust_2019}, are built upon probabilistic principles like Bayesian uncertainty quantification~\citep{feng_uncertainty-aware_2021}, or integrate sampling into the architecture and hence apply it also at inference time~\citep{chen_enhancing_2021, ioannidis_edge_2020, ioannidis_tensor_2020, luo_learning_2021}. \emph{Robust aggregation} defenses replace the message passing aggregation (typically mean) with a more robust equivalent such as a trimmed mean, median, or soft median~\citep{chen_understanding_2021, geisler_robustness_2021}. In relation to the trimmed mean, in this category we include also other related approaches that come with some guarantees based on their aggregation scheme~\citet{wang_provably_2020}.

\section{On adversarial training defenses}\label{sec:appendix:adversarial_training}

The most basic form of adversarial training for structure perturbations aims to solve:
\begin{equation}\label{eq:appendix:advtraining}
    \min_{\params} \max_{\adj' \in \pertmodel(\adj)} \loss(f_{\params}(\adj', \feat), \lab)
\end{equation}

Similarly to \citep{tramer_adaptive_2020, athalye_obfuscated_2018, carlini_towards_2017}, we exclude defenses that build on adversarial training in our study for three reasons.

First, we observe that adversarial training requires knowing the clean \(\adj\). However, for poisoning, we would need to substitute \(\adj\) with an adversarially perturbed adjacency matrix \(\pertadj\). In this case, adversarial training aims to enforce adversarial generalization \(\adj' \in \pertmodel(\pertadj)\) for the adversarially perturbed adjacency matrix \(\pertadj\) -- potentially even reinforcing the poisoning attack.

Second, an adaptive poisoning attack on adversarial training is very expensive as we need to unfold many adversarial attacks for a single training. Thus, designing truly adaptive poisoning attacks requires a considerable amount of resources. \emph{Scaling} these attacks to such complicated training schemes is not the main objective of this work.

Third, adversarial training for structure perturbations on GNNs seems to be an unsolved question. So far, the robustness gains come from additional and orthogonal tricks such as self-training~\citep{xu_topology_2019}. Hence, adversarial training for structure perturbations requires an entire paper on its own.

\section{On defenses against feature perturbations}\label{sec:appendix:feature_perturbations}

As introduced in \autoref{sec:preliminaries}, attacks may perturb the adjacency matrix \(\adj\), the feature matrix \(\feat\), or both. However, during our survey we found that few defenses tackle feature perturbations. Similarly, 6 out of the 7 defenses chosen by us mainly based on general popularity turn out to not consciously defend against feature perturbations.

The only exception is SVD-GCN~\citep{entezari_all_2020}, which also applies its low-rank approximation to the binary feature matrix. However, the authors do not report robustness under feature-only attacks; instead, they only consider mixed structure and feature attacks found by Nettack. Given the strong bias of Nettack towards structure perturbations, we argue that their experimental results do not confirm feature robustness. Correspondingly, in preliminary experiments we were not able to achieve considerable robustness gains of SVD-GCN compared to an undefended GCN -- even with non-adaptive feature perturbations. If a non-adaptive attack is strong enough, there is not much merit in applying an adaptive attack.

To reiterate, due to the apparent scarcity of defenses apt against feature attacks, we decided to focus our efforts on structure attacks and defenses. However, new defenses considering feature perturbations should study robustness in the face of adaptive attacks -- similarly to our work. In the following, we give some important hints for adaptive attacks using feature perturbations. We leave attacks that jointly consider feature and structure perturbations for future work due to the manifold open challenges, e.g., balancing structure and feature perturbations in the budget quantity.

\textbf{Baseline.}
To gauge the robustness of defenses w.r.t. global attacks, we introduce the RAUC metric, which employs the accuracy of an MLP -- which is perfectly robust w.r.t. structure perturbations -- to determine the maximally sensible budget to include in the summary. As MLPs are however vulnerable to feature attacks, a different baseline model is required for this new setting. We propose to resolve this issue by using a label propagation approach, which is oblivious to the node features and hence perfectly robust w.r.t. feature perturbations.

\textbf{Perturbations.}
The formulation of the set of admissible perturbations depends on what modality the data represents, which may differ between node features and graph edges. Convenient choices for continuous features are l-p-norms; in other cases, more complicated formulations are more appropriate. Accordingly, one has to choose an appropriate constrained optimization scheme.

\section{Examined adversarial defenses}\label{sec:appendix:defenses}

In this section, we portray each defense and how we adapted the base attacks to each one. We refer to \autoref{tab:appendix:model_hyperparams} for the used hyperparameter values for each defense. We give the used attack parameters for a GCN below and refer to the provided code for the other defenses.

\textbf{GCN.}
We employ an undefended GCN~\citep{kipf_semi-supervised_2017} as our baseline. A GCN first adds self loops to the adjacency matrix \(\adj\) and subsequently applies GCN-normalization, thereby obtaining \(\adj' = (\mathbf{D} + \mathbf{I})^{-\frac12} (\adj + \mathbf{I}) (\mathbf{D} + \mathbf{I})^{-\frac12}\) with the diagonal degree matrix \(\mathbf{D} \in \mathbb{N}^{n \times n}\). Then, in each GCN layer it updates the hidden states \(\hid^{(l)} = \dropout(\sigma(\adj' \hid^{(l-1)} \weight^{(l-1)} + \bias^{(l-1)}))\) where \(\hid^{(0)} = \feat\). We use the non-linear ReLU activation for intermediate layers. Dropout is deactivated in the last layer and we refer to the output before softmax activation as logits. We use Adam~\citep{kingma_adam_2015} to learn the model's parameters.

\textbf{Attack.}
We do not require special tricks since the GCN is fully differentiable and does not come with defensive measures to consider. In fact, the off-the-shelf attacks we employ are tailored to a GCN. For PGD, we use \(E=200\) iterations, \(K=100\) samples, and a base learning rate of 0.1. For Meta-PGD, we only lower the base learning rate to 0.01 and add gradient clipping to 1 (w.r.t. global \(L_2\)-norm). For Metattack with SGD instead of Adam for training the GCN, we use an SGD learning rate of 1 and restrict the training to \(E_{\text{train}}=100\) epochs.

\subsection{Jaccard-GCN}\label{sec:appendix:defenses_jaccard_gcn}

\textbf{Defense.}
Additionally to a GCN, Jaccard-GCN~\citep{wu_adversarial_2019} preprocesses the adjacency matrix. It computes the Jaccard coefficient of the binarized features for the pair of nodes of every edge, i.e., \(\mathbf{J}_{ij} = \frac{\feat_i \feat_j}{\min \{ \feat_i + \feat_j, 1 \}}\). Then edges are dropped where \(\mathbf{J}_{ij} \leq \epsilon\).

\textbf{Adaptive attack.}
We do not need to adapt gradient-based attacks as the gradient is equal to zero for dropped edges. Straightforwardly, we adapt Nettack to only consider non-dropped edges. Analogously, we ignore these edges in the Greedy Brute Force attack for increased efficiency.

\subsection{SVD-GCN}\label{sec:appendix:defenses_svd_gcn}

\textbf{Defense.}
SVD-GCN~\citep{entezari_all_2020} preprocesses the adjacency matrix with a low-rank approximation (LRA) for a fixed rank \(r\), utilizing the Singular Value Decomposition (SVD) \(\adj = \mathbf{U}\mathbf{\Sigma}\mathbf{V}^\top \approx \mathbf{U}_r\mathbf{\Sigma}_r\mathbf{V}_r^\top = \adj_r\). Note that the LRA is performed on \(\adj\) before adding self-loops and GCN-normalization (see above). Thereafter, the dense \(\adj_r\) is passed to the GCN as usual. Since \(\adj\) is symmetric and positive semi-definite, we interchangeably refer to the singular values/vectors also as eigenvalues/eigenvectors. 

\textbf{Adaptive attack.}
Unfortunately, the process of determining the singular vectors \(\mathbf{U}_r\) and \(\mathbf{V}_r\) is highly susceptible to small perturbations, and so is its gradient. Thus, we circumvent the need of differentiating the LRA. 

We now explain the approach from a geometrical perspective. Each row of \(\adj\) (or interchangeably column as \(\adj\) is symmetric) is interpreted as coordinates of a high-dimensional point. The \(r\) most significant eigenvectors of \(\adj\) span an \(r\)-dimensional subspace, onto which the points are projected by the LRA. Adding or removing an adversarial edge \((i,j)\) corresponds to moving the point \(\adj_i\) along dimension \(j\), i.e., \(\adj_i \pm \mathbf{e}_j\) (vice-versa for \(\adj_j\)). As hinted at in \autoref{sec:methodology}, the \(r\) most significant eigenvectors of \(\adj\) turn out to usually have few large components. Thus, the relevant subspace is mostly aligned with only few dimensions. 

Changes along the highest-valued eigenvectors are consequently preserved by LRA. To quantify how much exactly such a movement along a dimension \(j\), i.e., \(\mathbf{e}_j\), is preserved, we project the movement itself onto the subspace and extract the projected vector's \(j\)-th component. More formally, we denote the projection matrix onto the subspace as \(\mathbf{P} = \sum_{k=0}^r \mathbf{v}_k \mathbf{v}_k^T\) where \(\mathbf{v}_k\) are the eigenvectors of \(\adj\). We now score each dimension \(j\) with \((\mathbf{P} \mathbf{e}_j)_j = \mathbf{P}_{jj}\). Since the adjacency matrix is symmetric and rows and columns are hence exchangeable, we then symmetrize the scores \(\weight_{ij} = \sfrac{(\mathbf{P}_{ii} + \mathbf{P}_{jj})}{2}\).

Finally, we decompose the perturbed adjacency matrix \(\pertadj = \adj + \diffadj\) and, thus, only need gradients for \(\diffadj\). Using the approach sketched above, we now replace \(\operatorname{LRA}(\adj + \diffadj) \approx \operatorname{LRA}(\adj) + \diffadj \circ \weight\).

The weights \(\weight\) can also be incorporated into the Greedy Brute Force attack by dropping edges with weight \(< 0.2\) and, for efficient early stopping, sort edges to try in order of descending weight. Similarly, Nettack's score function \(s_\text{struct}(i,j)\) -- which attains positive and negative values, while \(\weight\) is positive -- can be wrapped to \(s'_\text{struct}(i,j) = \log (\exp(s_\text{struct}(i,j)) \circ \weight) = s_\text{struct}(i,j) + \log \weight\).

Note that we assume that the direction of the eigenvectors remains roughly equal after perturbing the adjacency matrix. In practice, we find this assumption to be true. Intuitively, a change along the dominant eigenvectors should even reinforce their significance.

\subsection{RGCN}

\textbf{Defense.}
The implementations of R(obust)GCN provided by the authors\footnote{\url{https://github.com/ZW-ZHANG/RobustGCN}} and in the widespread DeepRobust~\citep{li_deeprobust_2020} library\footnote{\url{https://github.com/DSE-MSU/DeepRobust} \label{foot:appendix:deeprobust}} are both consistent, but diverge slightly from the paper~\citep{zhu_robust_2019}. We use and now present RGCN according to those reference implementations. Principally, RGCN models the hidden states as Gaussian vectors with diagonal variance instead of sharp vectors. In addition to GCN's \(\adj'\), a second \(\adj'' = (\mathbf{D} + \mathbf{I})^{-1} (\adj + \mathbf{I}) (\mathbf{D} + \mathbf{I})^{-1}\) is prepared to propagate the variances. The mean and variance of this hidden Gaussian distribution are initialized as \(\mathbf{M}^{(0)} = \mathbf{V}^{(0)} = \feat\). Each layer first computes an intermediate distributions given by \(\hat{\mathbf{M}}^{(l)} = \elu(\dropout(\mathbf{M}^{(l-1)}) \weight_M^{(l-1)})\) and \(\hat{\mathbf{V}}^{(l)} = \relu(\dropout(\mathbf{V}^{(l-1)}) \weight_V^{(l-1)})\). Then, attention coefficients \(\bm{\alpha}^{(l)} = e^{-\gamma \hat{\mathbf{V}}^{(l)}}\) are calculated with the aim to subdue high-variance dimensions (where exponentiation is element-wise and \(\gamma\) is a hyperparameter). The final distributions are obtained with \(\mathbf{M}^{(l)} = \adj' \hat{\mathbf{M}}'^{(l)} \circ \bm{\alpha}^{(l)}\). Note the absence of bias terms. After the last layer, point estimates are sampled from the distributions via the reparameterization trick, i.e., scalars are sampled from a standard Gaussian and arranged in a matrix \(\mathbf{R}\). These samples are then used to obtain the logits via \(\mathbf{M}^{(L)} + \mathbf{R} \circ (\mathbf{V}^{(L)} + \epsilon)^{\frac12}\) (where the square root applies element-wise and \(\epsilon\) is a hyperparameter). Adam is the default optimizer. The loss is extended with the regularizer \(\beta \sum_i \operatorname{KL}(\mathcal{N}(\hat{\mathbf{M}}^{(1)}_i, \operatorname{diag}(\hat{\mathbf{V}}^{(1)}_i)) \| \mathcal{N}(\mathbf{0}, \mathbf{I}))\) (where \(\beta\) is a hyperparameter).

\textbf{Adaptive attack.}
A direct gradient attack suffices for a strong adaptive attack. Only when unrolling the training procedure for Metattack and Meta-PGD, we increase hyperparameter \(\epsilon\) from \(10^{-8}\) to  \(10^{-2}\) to retain numerical stability.

\subsection{ProGNN}

\textbf{Defense.}
We use and present Pro(perty)GNN~\citep{jin_graph_2020} exactly following the implementation provided by the authors in their DeepRobust~\citep{li_deeprobust_2020} library\textsuperscript{\ref{foot:appendix:deeprobust}}. ProGNN learns an alternative adjacency matrix \(\mathbf{S}\) that is initialized with \(\mathbf{A}\). A regular GCN -- which, as usual, adds self-loops and applies GCN-normalization -- is trained using \(\mathbf{S}\), which is simultaneously updated in every \(\tau\)-th epoch. For that, first a gradient descent step is performed on \(\mathbf{S}\) with learning rate \(\eta\) and momentum \(\mu\) towards minimizing the principal training loss alongside two regularizers that measure deviation \(\beta_1 \| \mathbf{S} - \adj \|_F^2\) and feature smoothness \(\frac{\beta_2}{2} \sum_{i,j} \mathbf{S}_{ij} \| \frac{\feat_i}{\sqrt{d_i}} - \frac{\feat_j}{\sqrt{d_j}} \|^2\) (where \(d_i = \sum_j \mathbf{S}_{ij} + 10^{-3}\)). Next, the singular value decomposition \(\mathbf{U} \mathbf{\Sigma} \mathbf{V}^T\) of the updated \(\mathbf{S}\) is computed, and \(\mathbf{S}\) is again updated to be \(\mathbf{U} \max(0, \mathbf{\Sigma} - \eta \beta_3) \mathbf{V}^T\) to promote low-rankness. Thereafter, \(\mathbf{S}\) is again updated to be \(\operatorname{sgn}(\mathbf{S}) \circ \max(0, |\mathbf{S}| - \eta \beta_4)\) to promote sparsity. Finally, the epoch's resulting \(\mathbf{S}\) is obtained by clamping its elements between 0 and 1.

\textbf{Adaptive attack.}
Designing an adaptive attack for ProGNN proved to be a challenging endeavor. We describe the collection of tricks in \autoref{sec:methodology}'s Example 2.

\subsection{GNNGuard}

\textbf{Defense.}
We closely follow the authors' implementation\footnote{\url{https://github.com/mims-harvard/GNNGuard}} as it deviates from the formal definitions in the paper~\citep{zhang_gnnguard_2020}. GNNGuard adopts a regular GCN and, before each layer, it adaptively weights down alleged adversarial edges. Thus, each layer has a unique propagation matrix \(\adj^{(l)}\) that is used instead of \(\adj'\).

GNNGuard's rule-based edge reweighting can be clustered into four consecutive steps: (1) the edges are reweighted based on the pair-wise cosine similarity \(\mathbf{C}^{(l)}_{ij} = \frac{\hid^{(l-1)}_i \cdot \hid^{(l-1)}_j}{\|\hid^{(l-1)}_i\| \|\hid^{(l-1)}_j\|}\) according to  \(\mathbf{S}^{(l)} = \adj \circ \mathbf{C}^{(l)} \circ \mathbb{I}[\mathbf{C}^{(l)} \ge 0.1]\), where edges with too dissimilar node embeddings are removed (see Iverson bracket \(\mathbb{I}[\mathbf{C}^{(l)} \ge 0.1]\)). Then, (2) the matrix is rescaled \(\bm{\Gamma}^{(l)}_{ij} = \sfrac{\mathbf{S}^{(l)}_{ij}}{\mathbf{s}^{(l)}_i}\) with \(\mathbf{s}^{(l)}_i = \sum_j \mathbf{S}^{(l)}_{ij}\) For stability, if \(\mathbf{s}^{(l)}_i < \epsilon\), \(\mathbf{s}^{(l)}_i\) is set to 1 (here \(\epsilon\) is a small constant). Next, (3) self-loops are added and \(\bm{\Gamma}^{(l)}\) is non-linarily transformed according to \(\hat{\bm{\Gamma}}^{(l)} = \exp_{\neq 0}(\bm{\Gamma}^{(l)} + \operatorname{diag} \sfrac{1}{1 + \mathbf{d}^{(l)}})\), where \(\exp_{\neq 0}\) only operates on nonzero elements and \(\mathbf{d}^{(l)}_i = \| \bm{\Gamma}^{(l)}_i \|_0\) is the row-wise number of nonzero entries. Last, (4) the result is smoothed over the layers with \(\bm{\Omega}^{(l)} = \sigma(\rho) \bm{\Omega}^{(l-1)} + (1-\sigma(\rho)) \hat{\bm{\Gamma}}^{(l)}\) with learnable parameter \(\rho\) and sigmoid function \(\sigma(\cdot)\).

The resulting reweighted adjacency matrix \(\bm{\Omega}^{(l)}\) is then GCN-normalized (without adding self-loops) and passed on to a GCN layer. Note that steps (1) to (3) are excluded from back-propagation during training. When comparing with the GNNGuard paper, one notices that among other deviations, we have omitted learnable edge pruning because it is disabled in the reference implementation.

\textbf{Adaptive attack.}
The hyperparameter \(\epsilon\) must be increased from \(10^{-6}\) to \(10^{-2}\) during the attack to retain numerical stability. In contrast to the reference implementation but as stated above, it is important to place the hard filtering step \(\mathbb{I}[\mathbf{C}^{(l)} \ge 0.1]\) for \(\mathbf{S}^{(l)}\) s.t. the gradient calculation w.r.t. \(\adj\) is not suppressed for these entries.

\subsection{GRAND}

\textbf{Defense.}
The Graph Random Neural Network (GRAND)~\citep{feng_graph_2021} model is the only defense from our selection that is not based on a GCN. First, \(\adj\) is endowed with self-loops and GCN-normalized to obtain \(\adj'\). Also, each row of \(\feat\) is \(l_1\)-normalized, yielding \(\feat'\). Next, rows from \(\feat'\) are randomly dropped with probability \(\delta\) during training to generate a random augmentation, and \(\feat'\) is scaled by \(1 - \delta\) during inference to compensate, thereby obtaining \(\hat{\feat}\). Those preprocessed node features are then propagated multiple times along the graph to get \(\overline{\feat} = \frac{1}{K+1} \sum_{k=0}^{K} \adj'^k \hat{\feat}\). Finally, dropout is applied once to \(\overline{\feat}\), and the result is plugged into a 2-layer MLP with dropout and ReLU activation to obtain class probabilities \(\mathbf{Z}\). The authors also propose an alternative architecture using a GCN instead of an MLP, however, we do not explore this option since the MLP version is superior according to their own results.

GRAND is trained with Adam. The training loss comprises the mean of the cross-entropy losses of \(S\) model evaluations, thereby incorporating multiple random augmentations. Additionally, a consistency regularizer is added to enforce similar class probabilities across all evaluations. More formally, first the probabilities are averaged across all evaluations: \(\overline{\mathbf{Z}} = \frac{1}{S} \sum_{s=1}^S \mathbf{Z}^{(s)}\). Next, each node's categorical distribution is sharpened according to a temperature hyperparameter \(T\), i.e., \(\overline{\mathbf{Z}}'_{ij} = \sfrac{\overline{\mathbf{Z}}_{ij}^{\frac{1}{T}}}{\sum_c \overline{\mathbf{Z}}_{ic}^{\frac{1}{T}}}\). The final regularizer penalizes the distance between the class probabilities and the sharpened averaged distributions, namely \(\frac{\beta}{S} \sum_{s=1}^S \| \mathbf{Z}^{(s)} - \overline{\mathbf{Z}}' \|_F^2\).

\textbf{Adaptive attack.}
When unrolling the training procedure for Metattack and Meta-PGD, to reduce the memory footprint, we reduce the number of random augmentations per epoch to 1, and we use a manual gradient calculation for the propagation operation. We also initialize Meta-PGD with a strong perturbation found by Meta-PGD on ProGNN. Otherwise, the attack has issues finding a perturbation with high loss; it presumably stalls in a local optimum. It is surprising that ``only'' initializing from GCN instead of ProGNN does not give a satisfyingly strong attack. Finally, we use the same random seed for every iteration of Metattack and Meta-PGD, as otherwise the constantly changing random graph augmentations make the optimization very noisy.

\subsection{Soft-Median-GDC}

\textbf{Defense.}
The Soft-Median-GDC~\citep{geisler_robustness_2021} deviates in two ways from a GCN: (1) it uses Personalized Page Rank (PPR) with restart probability \(\alpha=0.15\) to further preprocess the adjacency matrix after adding self-loops and applying GCN-normalization. The result is then sparsified using a row-wise top-\(k\) operation (\(k=64\)). (2) the message passing aggregation is replaced with a robust estimator called Soft-Median. From the perspective of node \(i\), a GCN uses the message passing aggregation \(\hid^{(l)}_i = \adj_i \hid^{(l-1)}\) which can be interpreted as a weighted mean/sum. In Soft-Median-GDC, the ``weights'' \(\adj_i\) are replaced with a scaled version of \(\adj_i \circ \text{softmax}\left(-\sfrac{\mathbf{c}}{T\sqrt{d}} \right)\). Here the vector \(\mathbf{c}\) denotes the distance between hidden embedding of a neighboring node to the neighborhood-specific weighted dimension-wise median: \(\mathbf{c}_{i} = \|\operatorname{Median}(\adj_i, \hid^{(l-1)}) - \hid^{(l-1)}_i\|\). To keep the scale, these weights are scaled s.t. they sum up to \(\sum \adj_i\).

\textbf{Adaptive attack.}
During gradient-based attacks, we adjust the \(\mathbf{c}\) of every node s.t. it now captures the distance to all other nodes, not only neighbors. This of course modifies the values of \(\mathbf{c}\), but is necessary to obtain a nonzero gradient w.r.t. to all candidate edges. We initialize PGD with a strong perturbation found by a similar attack on GCN, and initialize Meta-PGD with a perturbation from a similar attack on ProGNN (as with GRAND, using an attack against GCN as a base would be insufficient here).

\section{Evaluation of adaptive attacks}\label{sec:appendix:adaptive_attacks}

\begin{figure}[h]
    \centering
    \captionsetup[subfigure]{justification=centering}
    {
        \captionsetup[subfigure]{margin={1.4cm,0cm}}
        \subcaptionbox{Global, Poisoning}{\includegraphics[scale=0.64]{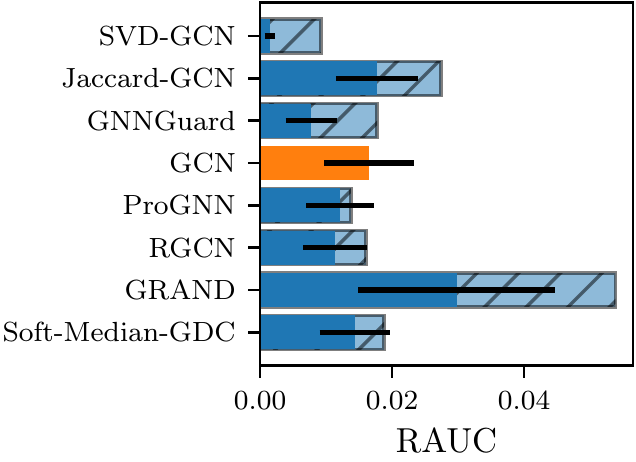}}
    }%
    \hfill
    \subcaptionbox{Global, Evasion}{\includegraphics[scale=0.64]{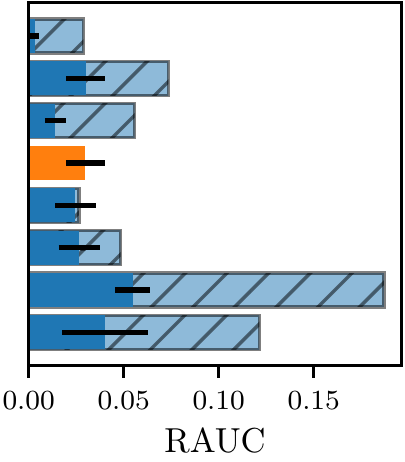}}
    \hfill
    \subcaptionbox{Local, Poisoning}{\includegraphics[scale=0.64]{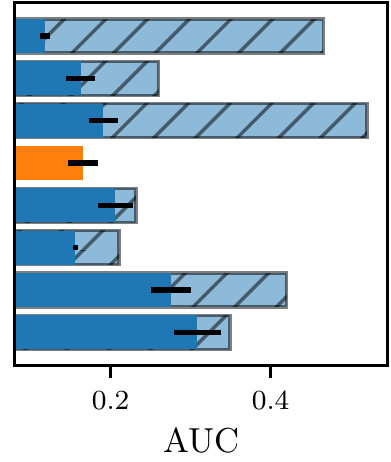}}
    \hfill
    {
        \captionsetup[subfigure]{margin={0cm,1.6cm}}
        \subcaptionbox{Local, Evasion}{\includegraphics[scale=0.64]{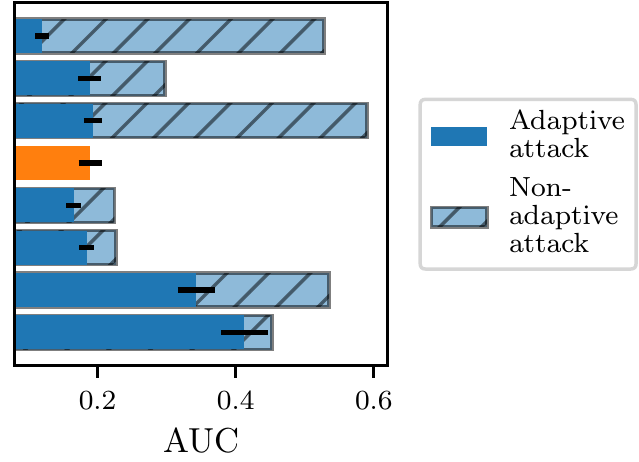}}
    }%
    \caption{Variant of \autoref{fig:improvement_adaptive_auc} for Citeseer.}
    \label{fig:appendix:improvement_adaptive_auc_citeseer}
\end{figure}

\begin{figure}[p]
    \centering

    \hspace*{0.85cm} \includegraphics[scale=0.8]{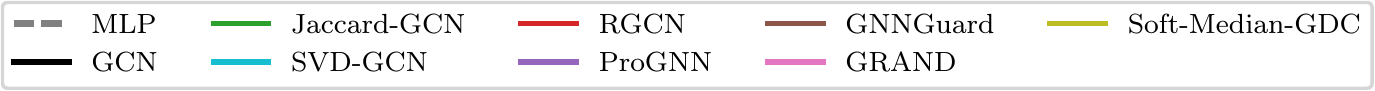}

    \vspace{0.2cm}
    \begin{subfigure}{0.48\linewidth}
        \includegraphics[width=\linewidth]{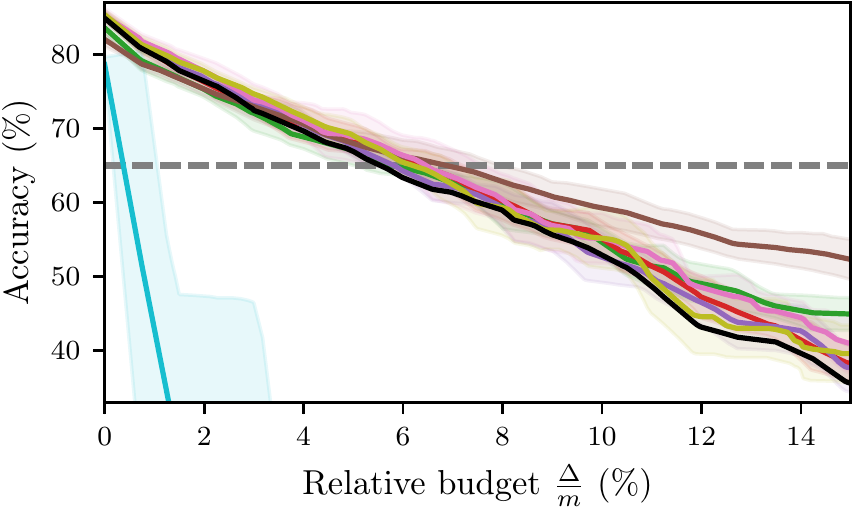}
        \captionsetup{margin={0.8cm,0cm}} \caption{Cora~ML, Poisoning}
    \end{subfigure}
    \hfill
    \begin{subfigure}{0.48\linewidth}
        \includegraphics[width=\linewidth]{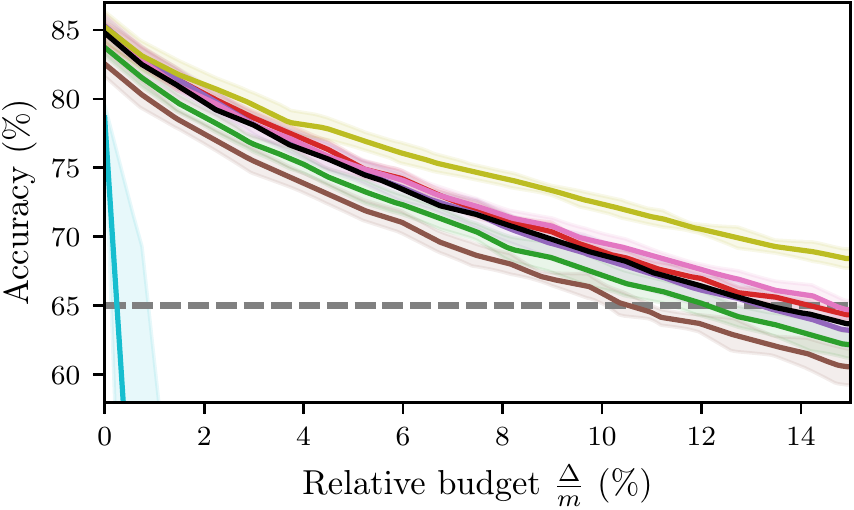}
        \captionsetup{margin={0.8cm,0cm}} \caption{Cora~ML, Evasion}
    \end{subfigure}

    \vspace{0.3cm}
    \begin{subfigure}{0.48\linewidth}
        \includegraphics[width=\linewidth]{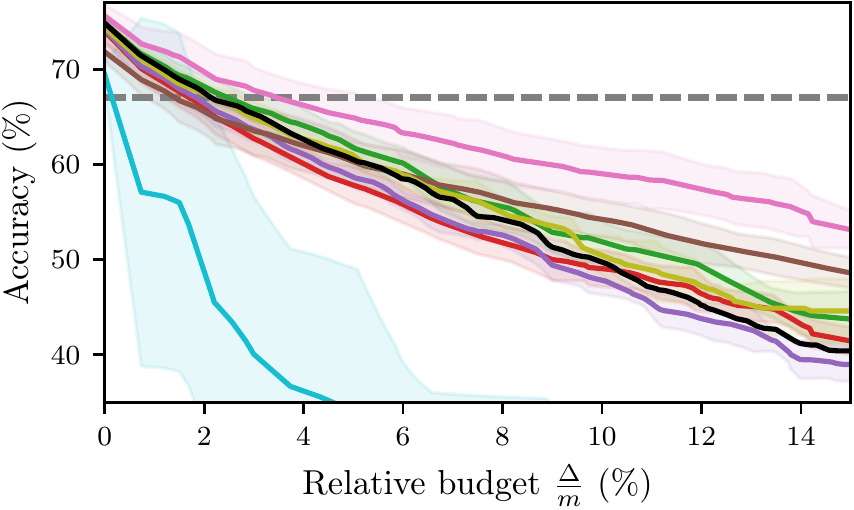}
        \captionsetup{margin={0.8cm,0cm}} \caption{Citeseer, Poisoning}
    \end{subfigure}
    \hfill
    \begin{subfigure}{0.48\linewidth}
        \includegraphics[width=\linewidth]{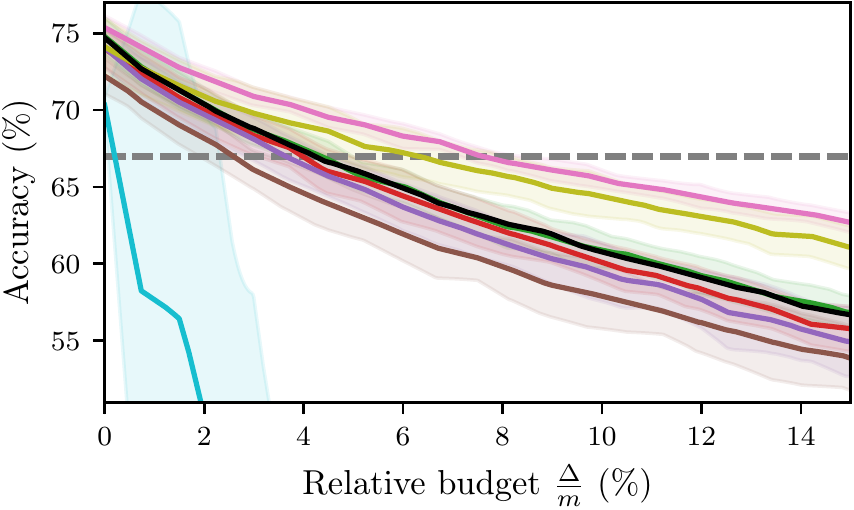}
        \captionsetup{margin={0.8cm,0cm}} \caption{Citeseer, Evasion}
    \end{subfigure}

    \caption{Absolute variant of \autoref{fig:global_mr_envelope}, showing relative budgets up to 15\%.}
    \label{fig:appendix:abs_global_mr_envelope}

    \vspace{1.8\floatsep}

    \hspace*{0.85cm} \includegraphics[scale=0.8]{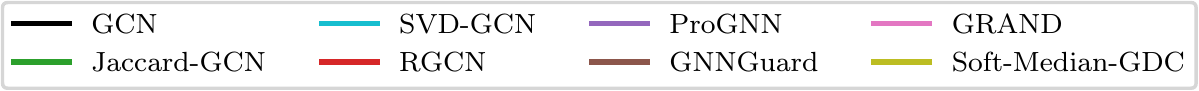}

    \vspace{0.2cm}
    \begin{subfigure}{0.48\linewidth}
        \includegraphics[width=\linewidth]{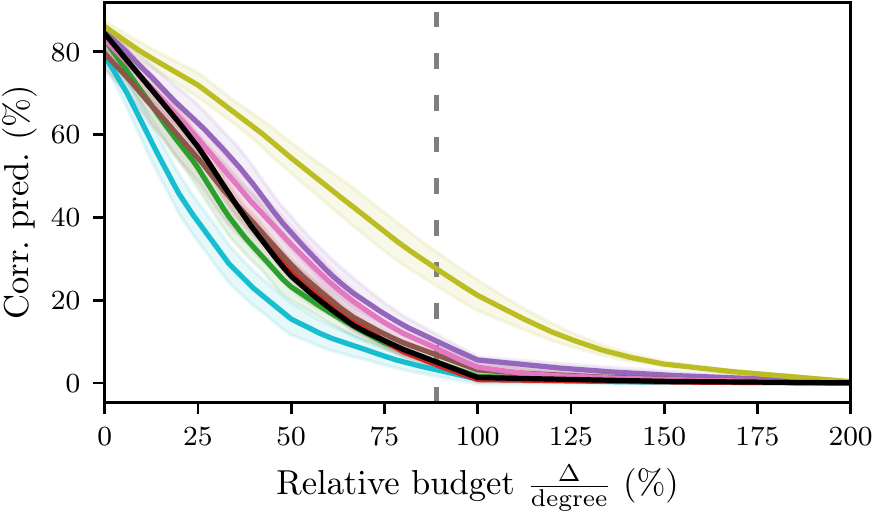}
        \captionsetup{margin={0.8cm,0cm}} \caption{Cora~ML, Poisoning}
    \end{subfigure}
    \hfill
    \begin{subfigure}{0.48\linewidth}
        \includegraphics[width=\linewidth]{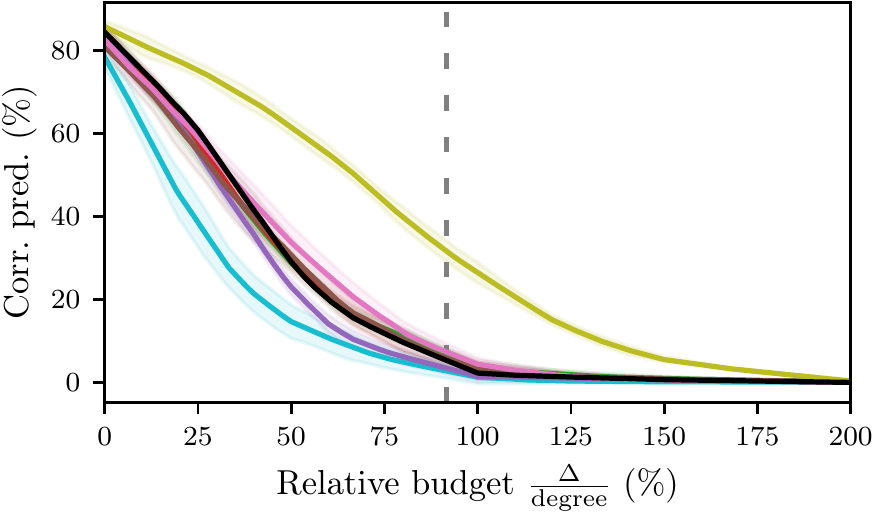}
        \captionsetup{margin={0.8cm,0cm}} \caption{Cora~ML, Evasion}
    \end{subfigure}

    \vspace{0.3cm}
    \begin{subfigure}{0.48\linewidth}
        \includegraphics[width=\linewidth]{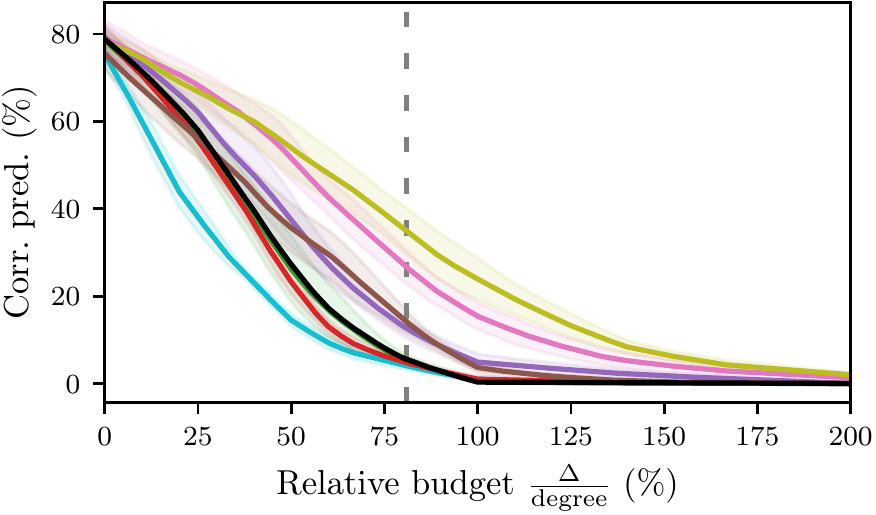}
        \captionsetup{margin={0.8cm,0cm}} \caption{Citeseer, Poisoning}
    \end{subfigure}
    \hfill
    \begin{subfigure}{0.48\linewidth}
        \includegraphics[width=\linewidth]{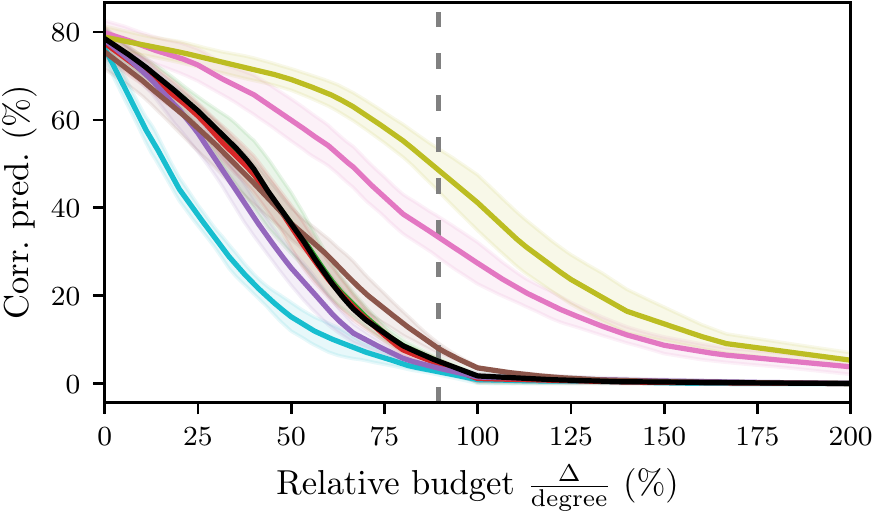}
        \captionsetup{margin={0.8cm,0cm}} \caption{Citeseer, Evasion}
    \end{subfigure}

    \caption{Absolute variant of \autoref{fig:local_mr_envelope}, showing relative budgets up to 200\%.}
    \label{fig:appendix:abs_local_breakage_envelope}
\end{figure}

In \autoref{tab:appendix:datasets}, we summarize the variants of the datasets we use, both of which we have precisely extracted from Nettack's code\footnote{\url{https://github.com/danielzuegner/nettack}}. In \autoref{fig:appendix:improvement_adaptive_auc_citeseer}, we complement \autoref{fig:improvement_adaptive_auc} and compare the (R)AUC of all defenses on Citeseer. The robustness estimates for the defenses on Citeseer are also much lower as originally reported. For completeness, we give absolute envelope curve plots for all settings and datasets as well as for higher budgets in \autoref{fig:appendix:abs_global_mr_envelope} and \autoref{fig:appendix:abs_local_breakage_envelope} (compare with \autoref{fig:global_mr_envelope} and \autoref{fig:local_mr_envelope}).

\begin{table}[h]
    \centering
    \caption{Statistics of the datasets we used. We measure homophily as the fraction of edges which connect nodes of the same class. \label{tab:appendix:datasets}}
    \vspace{0.5em}
    \resizebox{0.9\linewidth}{!}{
        \begin{tabular}{lllllll}
            \toprule
            Dataset & Nodes & Undirected Edges & Features & Classes & Avg. Degree & Homophily \\
            \midrule
            Cora ML \citep{bojchevski2018deep} & 2485 & 5069 & 1433 & 7 & 4.08 & 0.804 \\
            Citeseer \citep{giles1998citeseer} & 2110 & 3668 & 3703 & 6 & 3.477 & 0.736 \\
            \bottomrule
        \end{tabular}
    }
\end{table}

\newpage
\section{Ensemble transferability study}\label{sec:appendix:ensemble_transfer}

\begin{figure}[h]
    \centering
    {
        \captionsetup[subfigure]{margin={2.7cm,0cm}}
        \subcaptionbox{Poisoning}{\includegraphics[scale=0.67]{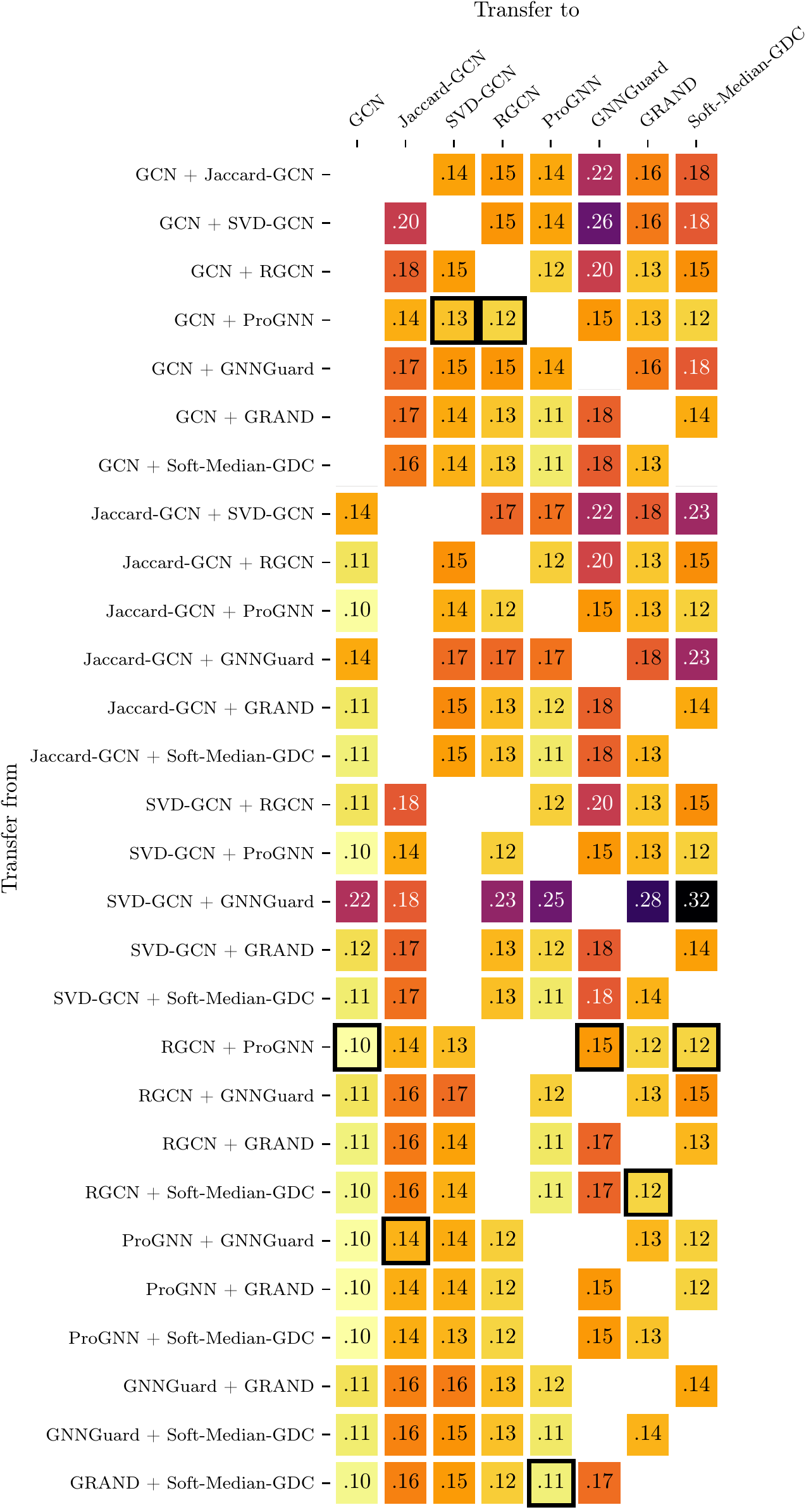}}
    }%
    \hspace{-0.2cm}
    {
        \captionsetup[subfigure]{margin={0cm,0.9cm}}
        \subcaptionbox{Evasion}{\includegraphics[scale=0.67]{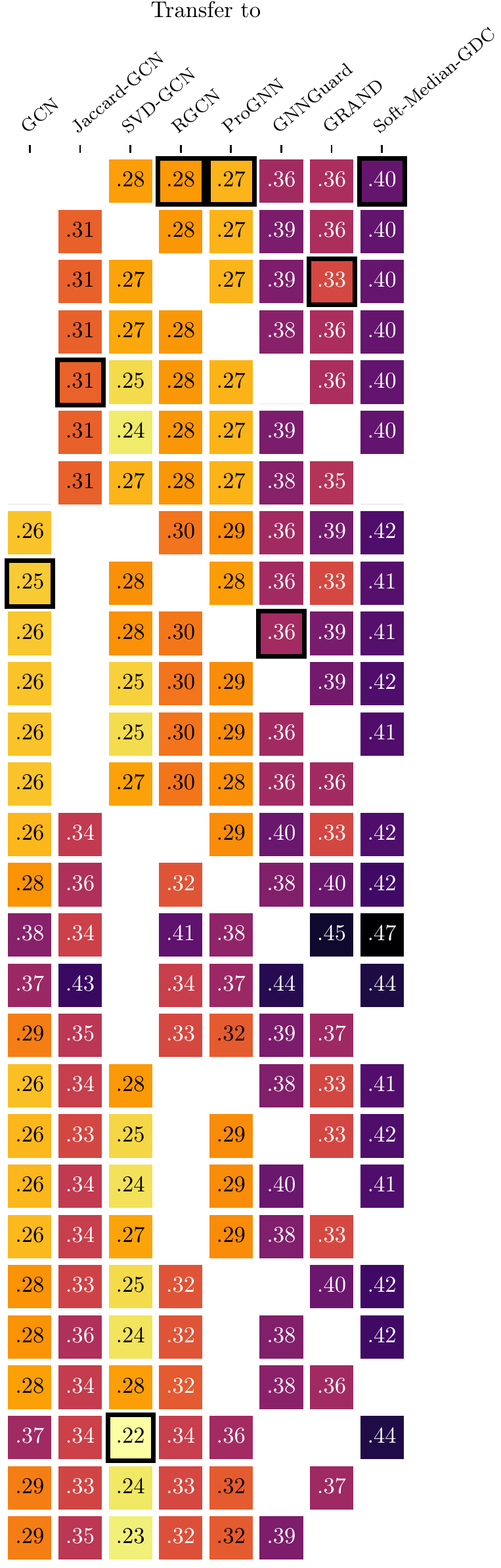}}
    }%
    \caption{Variant of \autoref{fig:transfer} with ensembles of models as attack transfer sources. The color maps are \emph{not} matched across (a) and (b) for improved readability.}
    \label{fig:appendix:ensemble_transfer}
\end{figure}

In \autoref{fig:transfer}, we transfer attacks found on an \emph{individual} model to other models. It is natural to also assess the strength of transfer attacks supplied by \emph{ensembles} of models. In \autoref{fig:appendix:ensemble_transfer}, we address this question for 2-ensembles. For poisoning, the combination of RGCN and ProGNN turns out to be (nearly) the strongest in all cases, which is reasonable since both already form strong individual transfer attacks as is evident in \autoref{fig:transfer}. For evasion, the differences are more subtle.

We also investigate 3-ensembles, but omit the plots due to their size. For poisoning, RGCN and ProGNN now combined with Soft-Median-GDC remain the strongest transfer source, yet the improvement over the 2-ensemble is marginal. For evasion, there is still no clear winner.

\section{GCN and defense hyperparameters: original vs. tuned for adaptive attacks}\label{sec:appendix:models_tuned_vs_original}

\begin{figure}[!b]
    \centering

    \includegraphics[scale=0.73]{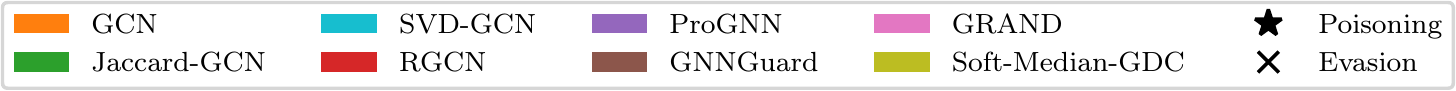}

    \vspace{0.15cm}
    \captionsetup[subfigure]{margin={1cm,0cm},skip=0.15cm}
    \subcaptionbox{Global, Cora~ML}{\includegraphics[scale=0.73]{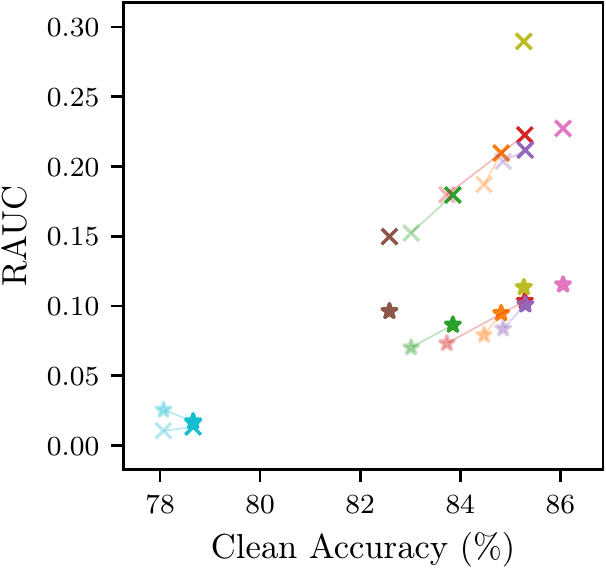}}
    \hspace{0.3cm}
    \captionsetup[subfigure]{margin={0.5cm,0cm}}
    \subcaptionbox{Global, Citeseer}{\includegraphics[scale=0.73]{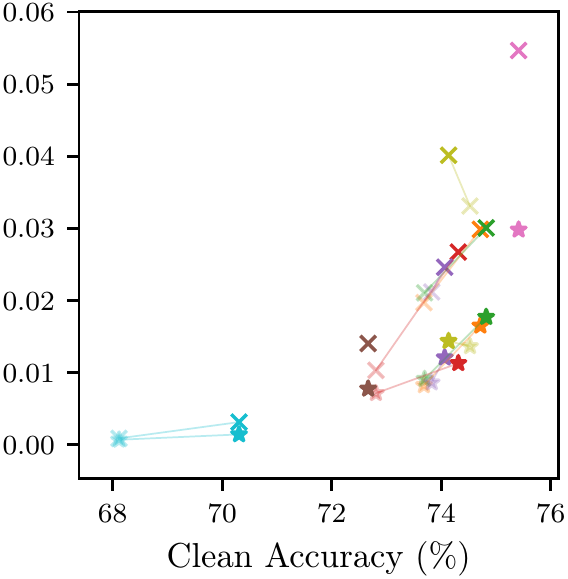}}

    \vspace{0.25cm}
    \captionsetup[subfigure]{margin={1cm,0cm}}
    \subcaptionbox{Local, Cora~ML}{\includegraphics[scale=0.73]{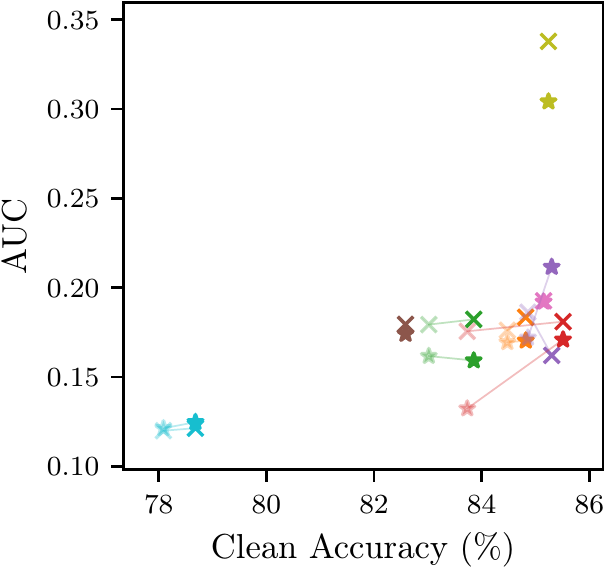}}
    \hspace{0.3cm}
    \captionsetup[subfigure]{margin={0.5cm,0cm}}
    \subcaptionbox{Local, Citeseer}{\includegraphics[scale=0.73]{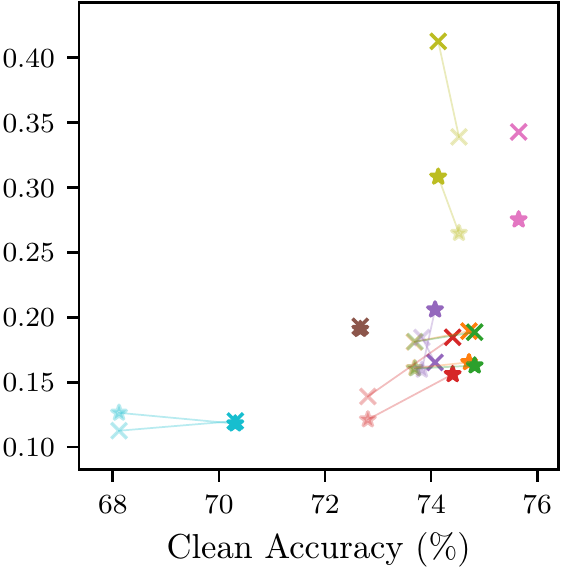}}

    \caption{Each defense's clean accuracy vs. (R)AUC values of the strongest attacks, akin to \autoref{fig:accuracy_robustness_tradeoff}. Muted (semi-transparent) colors represent untuned defenses (except for Soft-Median-GDC on Cora~ML and GNNGuard), solid colors denote tuned defenses, and lines connect the two. Our tuned defenses are almost always better than untuned variants w.r.t. both clean accuracy and robustness.}
    \label{fig:appendix:tuning_improvement}
\end{figure}

To allow for the fairest comparison possible, we tuned the hyperparameters for each model (including GCN) towards maximizing both clean accuracy and adversarial robustness on a single random data split. In \autoref{tab:appendix:model_hyperparams}, we list all hyperparameter configurations. While we cannot run an exhaustive search over all hyperparameter settings, we report substantial gains for most defenses and the GCN in \autoref{fig:appendix:tuning_improvement}. The only exceptions are GRAND, Soft-Median-GDC on Cora~ML, and GNNGuard. For GRAND, we do not report results for the default hyperparameters as they did not yield satisfactory clean accuracy. Moreover, for Soft-Median-GDC on Cora~ML and GNNGuard we were not able to substantially improve over the default hyperparameters.

For the GCN, tuning is important to ensure that we have a fair and equally-well tuned baseline. A GCN is the natural baseline since most defense methods propose slight modifications of a GCN or additional steps to improve the robustness. For the defenses, tuning is vital since they were most originally tuned w.r.t. non-adaptive attacks. In any case, the tuning should counterbalance slight variations in the setup.

As stated in the introduction, each attack only provides an upper bound for the actual adversarial robustness of a model (with fixed hyperparameters). A future attack of increased efficacy might lead to a tighter estimate. Thus, when we empirically compare the defenses to a GCN, we only compare upper bounds of the respective actual robustness. However, we attack the GCN with state-of-the-art approaches that were developed by multiple researchers specifically for a GCN. Even though we also tune the parameters of the adaptive attacks, we argue that the robustness estimate for a GCN is likely tighter than our robustness estimate for the defenses. In summary, the tuning of hyperparameters is necessary that we can fairly compare the robustness of multiple models, even though, we only compare upper bounds of the true robustness.

\begin{landscape}
\begin{table}[h]
\centering
\caption{GCN and defense hyperparameters.}
\label{tab:appendix:model_hyperparams}
\vspace{0.5em}
\resizebox{\linewidth}{!}{
\begin{tabular}{lll|lllll|llllllllllll}
\toprule
\multirow{3}{*}{GCN} & Tuned & & Hidden & Dropout & & & & Max epochs & Patience & LR & \(L_2\) reg. \\
\cmidrule{2-19}
& \(\times\) & & 1 \(\times\) 16 & 0.5 & & & & 3000 & 50 & 0.01 & 0.0005 \\
& \checkmark & & 1 \(\times\) 64 & 0.9 & & & & 3000 & 50 & 0.01 & 0.001 \\
\midrule \midrule
\multirow{3}{*}{Jaccard-GCN} & Tuned & & Hidden & Dropout & \(\epsilon\) & & & Max epochs & Patience & LR & \(L_2\) reg. \\
\cmidrule{2-19}
& \(\times\) & & 1 \(\times\) 16 & 0.5 & 0.0 & & & 3000 & 200 & 0.01 & 0.0005 \\
& \checkmark & & 1 \(\times\) 64 & 0.9 & 0.0 & & & 3000 & 50 & 0.01 & 0.001 \\
\midrule \midrule
\multirow{3}{*}{SVD-GCN} & Tuned & & Hidden & Dropout & Rank & & & Max epochs & Patience & LR & \(L_2\) reg. \\
\cmidrule{2-19}
& \(\times\) & & 1 \(\times\) 16 & 0.5 & 50 & & & 3000 & 200 & 0.01 & 0.0005 \\
& \checkmark & & 1 \(\times\) 64 & 0.9 & 50 & & & 3000 & 50 & 0.01 & 0.001 \\
\midrule \midrule
\multirow{3}{*}{RGCN} & Tuned & & Hidden & Dropout & \(\epsilon\) & \(\gamma\) & & Max epochs & Patience & LR & \(L_2\) reg. & \(\beta\) \\
\cmidrule{2-19}
& \(\times\) & & 1 \(\times\) 16 & 0.6 & 1e-8 & 1.0 & & 3000 & 50 & 0.01 & 0.0005 & 0.0005 \\
& \checkmark & & 1 \(\times\) 32 & 0.6 & 1e-8 & 1.0 & & 3000 & 50 & 0.01 & 0.01 & 0.0005 \\
\midrule \midrule
\multirow{5}{*}{ProGNN} & Tuned & & Hidden & Dropout & & & & Max epochs & Patience & LR & \(L_2\) reg. & \(\tau\) & \(\eta\) & \(\mu\) & \(\beta_1\) & \(\beta_2\) & \(\beta_3\) & \(\beta_4\) \\
\cmidrule{2-19}
& \multirow{2}{*}{\(\times\)} & Cora ML & 1 \(\times\) 16 & 0.5 & & & & 3000 & 50 & 0.01 & 0.0005 & 2 & 0.01 & 0.9 & 1.0 & 0.001 & 1.5 & 0.0005 \\
& & Citeseer & 1 \(\times\) 16 & 0.5 & & & & 3000 & 50 & 0.01 & 0.0005 & 2 & 0.01 & 0.9 & 1.0 & 0.0001 & 1.5 & 0.0005 \\
& \multirow{2}{*}{\checkmark} & Cora ML & 1 \(\times\) 16 & 0.5 & & & & 3000 & 50 & 0.01 & 0.0005 & 2 & 0.01 & 0.9 & 1.0 & 0.1 & 10.0 & 0.1 \\
& & Citeseer & 1 \(\times\) 16 & 0.5 & & & & 3000 & 50 & 0.01 & 0.0005 & 2 & 0.01 & 0.9 & 1.0 & 0.2 & 20.0 & 0.2 \\
\midrule \midrule
\multirow{2}{*}{GNNGuard} & Tuned & & Hidden & Dropout & Pruning & \(\epsilon\) & & Max epochs & Patience & LR & \(L_2\) reg. \\
\cmidrule{2-19}
& \(\times\) & & 1 \(\times\) 16 & 0.5 & \(\times\) & 1e-6 & & 81 & n/a & 0.01 & 0.0005 \\
\midrule \midrule
\multirow{3}{*}{GRAND} & Tuned & & Hidden & Dropout & \(\overline{\feat}\) dropout & \(\delta\) & \(K\) & Max epochs & Patience & LR & \(L_2\) reg. & \(S\) & \(\beta\) & \(T\) \\
\cmidrule{2-19}
& \multirow{2}{*}{\checkmark} & Cora ML & 1 \(\times\) 32 & 0.5 & 0.5 & 0.5 & 8 & 3000 & 50 & 0.05 & 0.0001 & 4 & 1.0 & 0.5 \\
& & Citeseer & 1 \(\times\) 32 & 0.2 & 0.0 & 0.5 & 2 & 3000 & 50 & 0.05 & 0.0005 & 2 & 0.7 & 0.3 \\
\midrule \midrule
\multirow{3}{*}{Soft-Median-GDC} & Tuned & & Hidden & Dropout & \(k\) & \(\alpha\) & \(T\) & Max epochs & Patience & LR & \(L_2\) reg. \\
\cmidrule{2-19}
& \(\times\) & & 1 \(\times\) 64 & 0.5 & 64 & 0.15 & 0.5 & 3000 & 50 & 0.01 & 0.001 \\
& \checkmark & Citeseer & 1 \(\times\) 64 & 0.5 & 64 & 0.25 & 0.5 & 3000 & 50 & 0.01 & 0.001 \\
\bottomrule
\end{tabular}
}
\end{table}
\end{landscape}

\section{Comparison of success of attack approaches}\label{sec:appendix:attack_success}

\begin{figure}[h]
    \centering

    \hspace*{2.1cm} \includegraphics[scale=0.8]{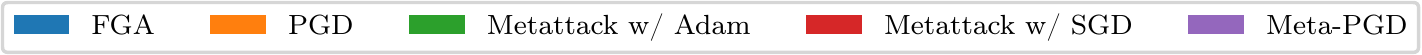}

    \vspace{0.1cm}
    {
        \captionsetup[subfigure]{margin={2.1cm,0cm}}
        \subcaptionbox{Cora~ML, Pois.}{\includegraphics[scale=0.8]{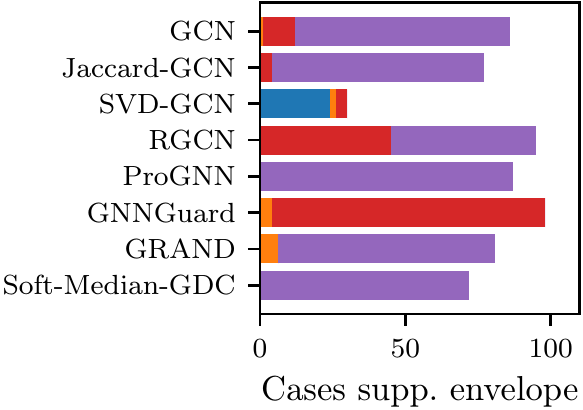}}
    }%
    \hfill
    \subcaptionbox{Citeseer,  Pois.}{\includegraphics[scale=0.8]{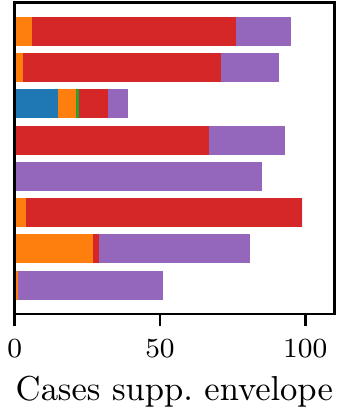}}
    \hfill
    \subcaptionbox{Cora~ML, Evas.}{\includegraphics[scale=0.8]{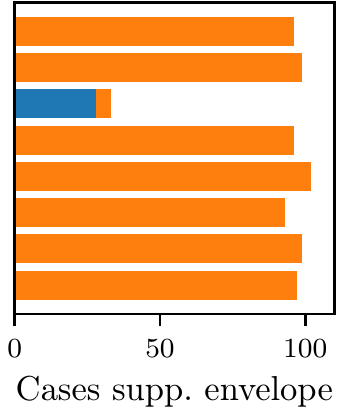}}
    \hfill
    \subcaptionbox{Citeseer, Evas.}{\includegraphics[scale=0.8]{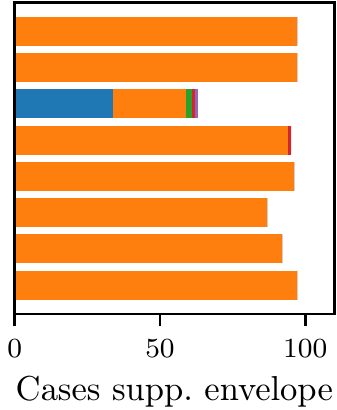}}

    \caption{Number of global attack attempts which support the envelope curve over all attack attempts, as introduced in \autoref{fig:rauc}. We observe that for evasion, PGD almost always yields the strongest attack, while for poisoning, either Metattack, Meta-PGD, or both dominate. Using Adam instead of SGD to train the defense nearly always worsens Metattack's performance.}
    \label{fig:appendix:global_envelope_support}

    \vspace{2\floatsep}

    \hfill \includegraphics[scale=0.8]{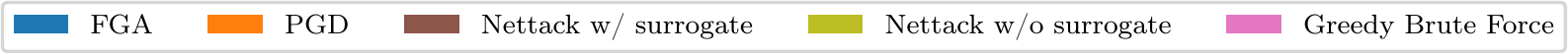}

    \vspace{0.1cm}
    {
        \captionsetup[subfigure]{margin={2.1cm,0cm}}
        \subcaptionbox{Cora~ML, Pois.}{\includegraphics[scale=0.8]{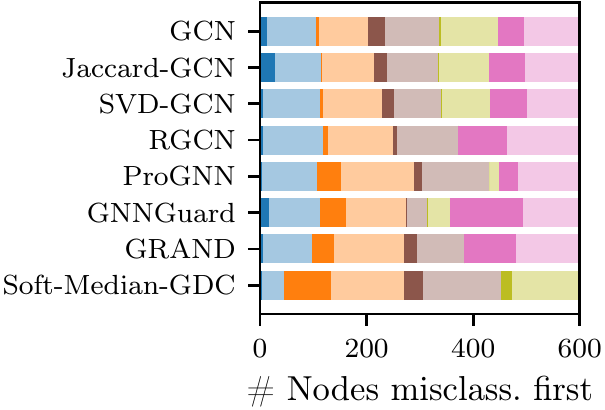}}
    }%
    \hfill
    \subcaptionbox{Citeseer,  Pois.}{\includegraphics[scale=0.8]{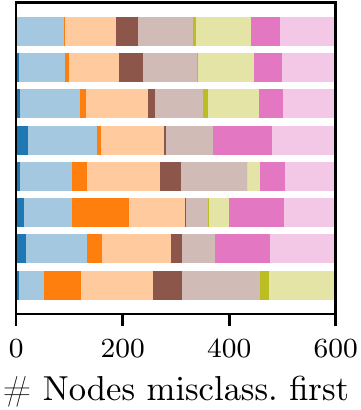}}
    \hfill
    \subcaptionbox{Cora~ML, Evas.}{\includegraphics[scale=0.8]{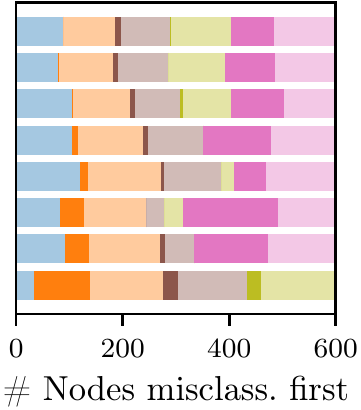}}
    \hfill
    \subcaptionbox{Citeseer, Evas.}{\includegraphics[scale=0.8]{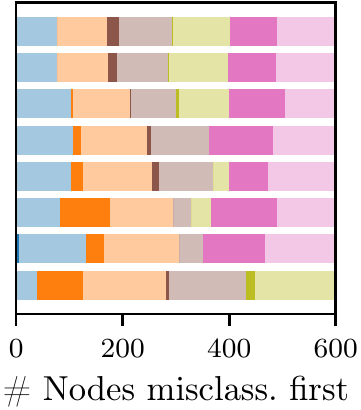}}

    \caption{Number of target nodes for which the respective local attack needs the least budget (among all attacks) to misclassify them. When multiple attacks achieve the same lowest budget, the target node is counted in parts towards each winning attack and drawn with a muted color. We observe that greedy brute force is often the strongest attack; only sometimes, PGD and Nettack beat it on some defenses, especially for poisoning. Using the defense's weights instead of a surrogate model for Nettack is rarely an improvement. Still, for the majority of target nodes, multiple attacks are equally strong in terms of achieving the same lowest budget (tie). We do not run the greedy brute force attack on Soft-Median-GDC due to the costly PPR calculation.}
    \label{fig:appendix:local_envelope_support}
\end{figure}

\begin{figure}[p]
    \centering

    \hspace*{2cm} \includegraphics[scale=0.8]{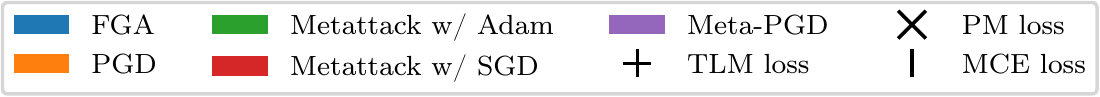}

    \vspace{0.2cm}
    \captionsetup[subfigure]{margin={1.8cm,0cm},skip=0.12cm}
    \subcaptionbox{Cora ML, Poisoning}{\includegraphics[scale=0.8]{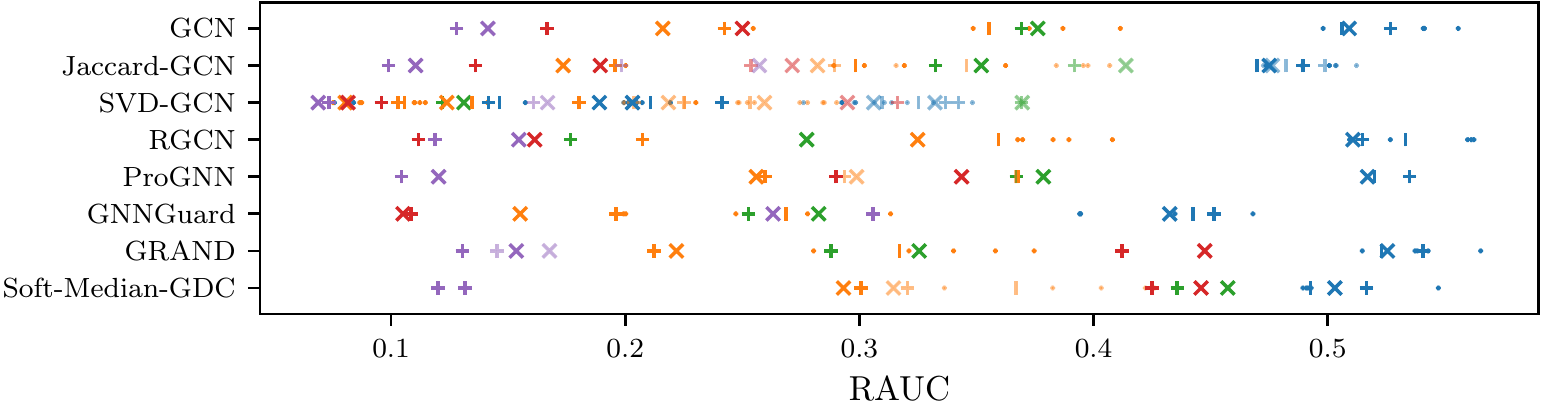}}

    \vspace{0.3cm}
    \subcaptionbox{Citeseer, Poisoning}{\includegraphics[scale=0.8]{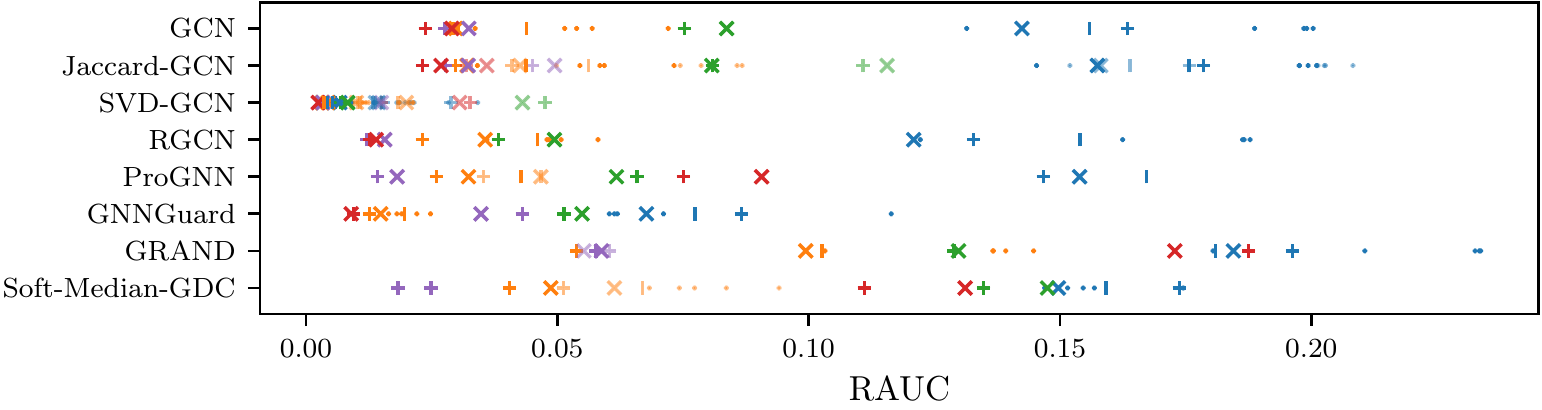}}

    \vspace{0.3cm}
    \subcaptionbox{Cora ML, Evasion}{\includegraphics[scale=0.8]{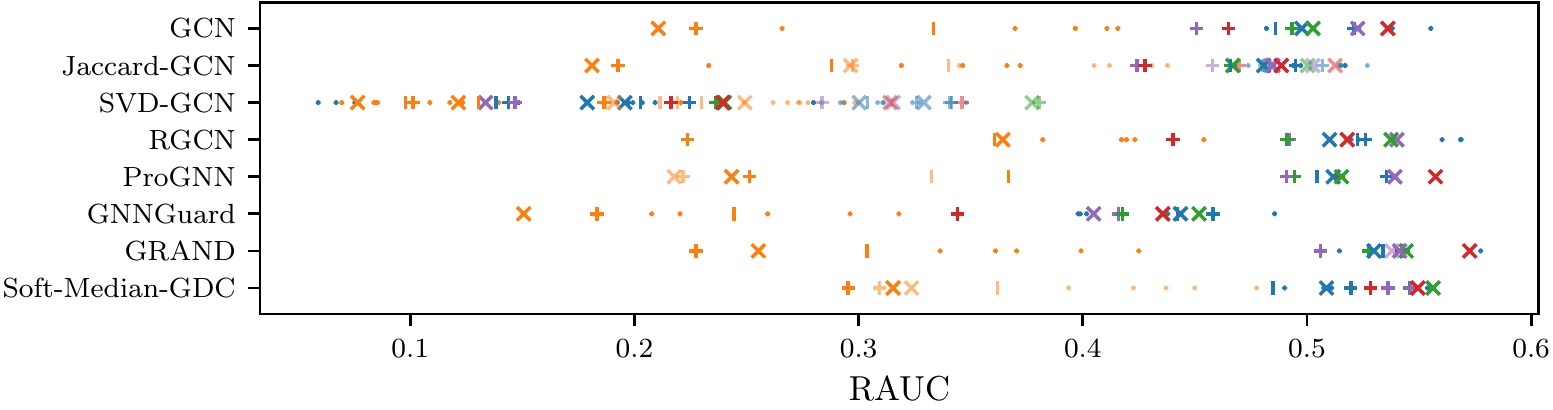}}

    \vspace{0.3cm}
    \subcaptionbox{Citeseer, Evasion}{\includegraphics[scale=0.8]{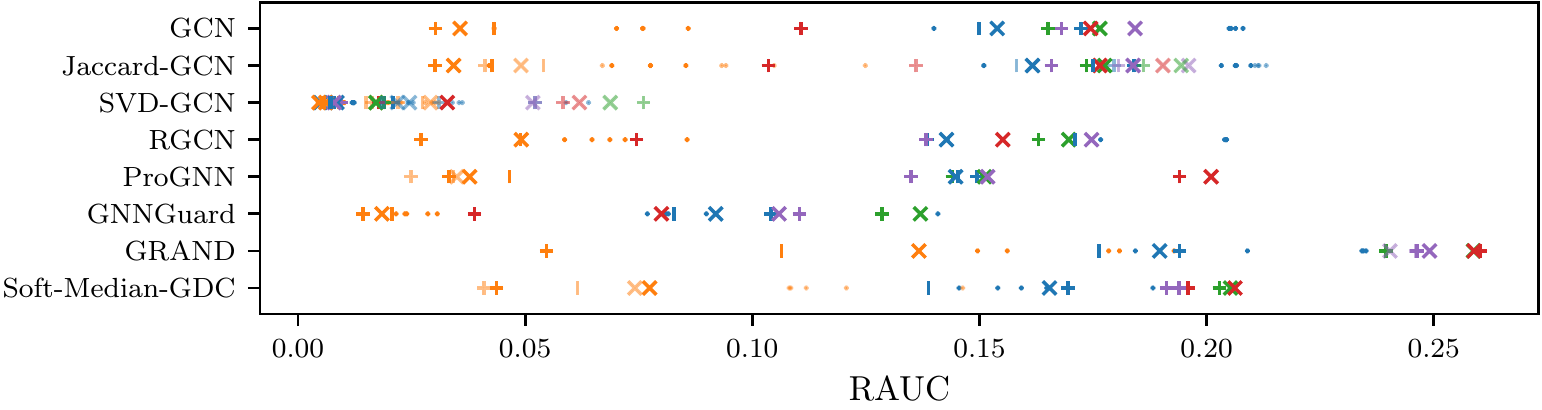}}

    \caption{The RAUC of every global attack we have conducted. Attacks are color-coded by principal technique, and markers indicate the attack loss. Muted colors represent attacks \emph{without} edge masking (Jaccard-GCN), our edge weighting trick (SVD-GCN), multiple PGD auxiliary models (ProGNN), Meta-PGD initialization from ProGNN and unlimited unrolled epochs (GRAND), and PGD initialization from GCN (Soft-Median-GDC). We observe that (1) the TLM and PM losses are superior in almost all cases; (2) PGD attacks are best for evasion while Metattack and Meta-PGD are unsuited; (3) Metattack with SGD and Meta-PGD are best for poisoning while Metattack w/ Adam even falls behind the surprisingly strong evasion-poisoning transfer; (4) FGA is weak for each defense apart from SVD-GCN; (5) the cited adaptions are beneficial as attacks with muted colors are worse; (6) a strong adaptive attack is necessary to reach a low RAUC.}
    \label{fig:appendix:global_method_comparison}
\end{figure}

\begin{figure}[p]
    \centering

    \hfill \includegraphics[scale=0.85]{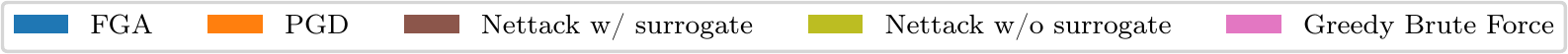}

    \vspace{0.2cm}
    \captionsetup[subfigure]{margin={1.8cm,0cm},skip=0.12cm}
    \subcaptionbox{Cora ML, Poisoning}{\includegraphics[scale=0.85]{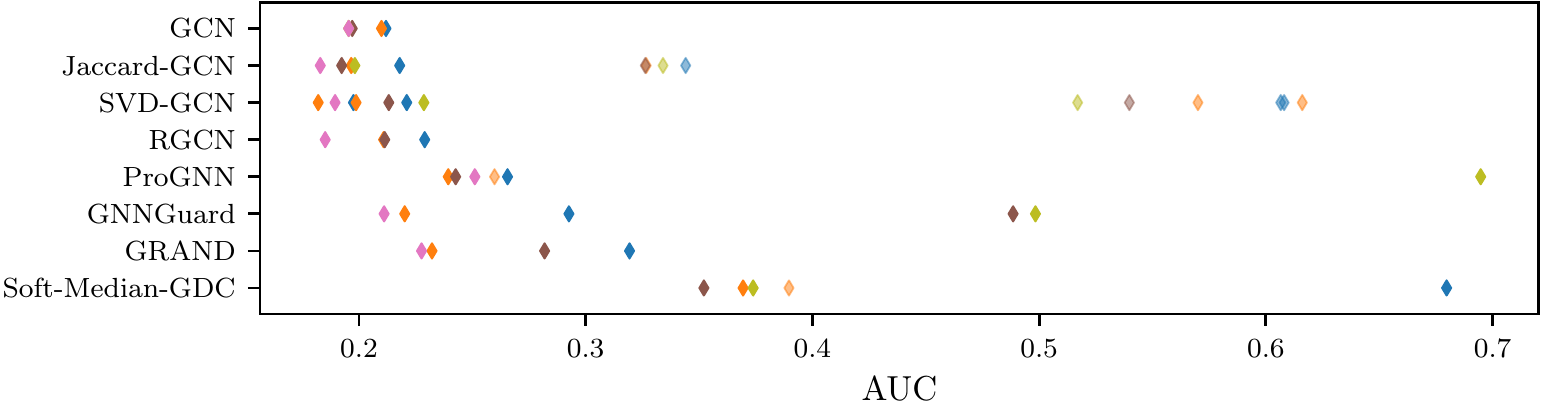}}

    \vspace{0.3cm}
    \subcaptionbox{Citeseer, Poisoning}{\includegraphics[scale=0.85]{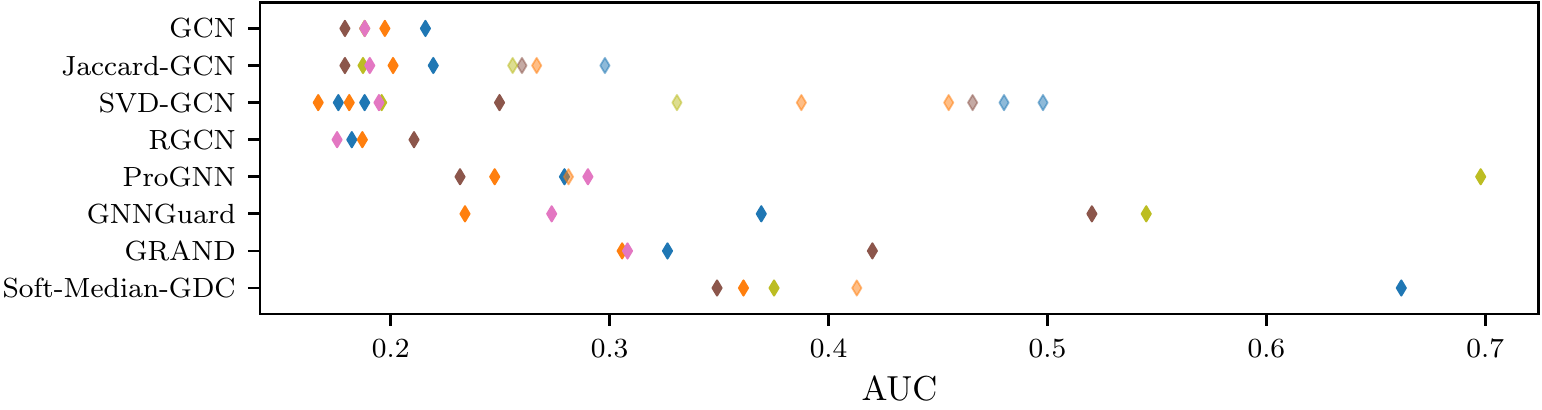}}

    \vspace{0.3cm}
    \subcaptionbox{Cora ML, Evasion}{\includegraphics[scale=0.85]{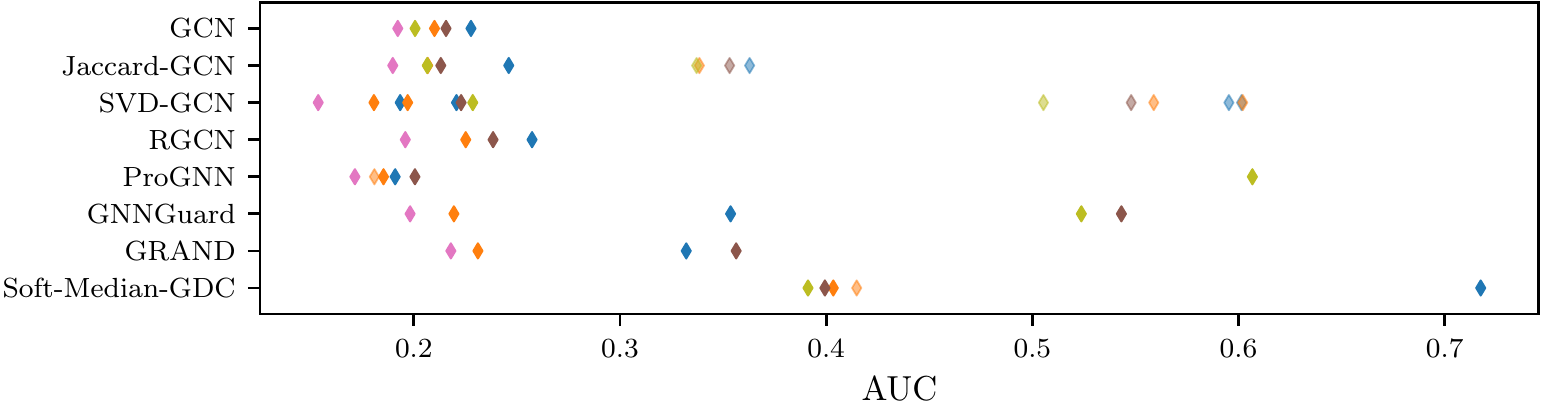}}

    \vspace{0.3cm}
    \subcaptionbox{Citeseer, Evasion}{\includegraphics[scale=0.85]{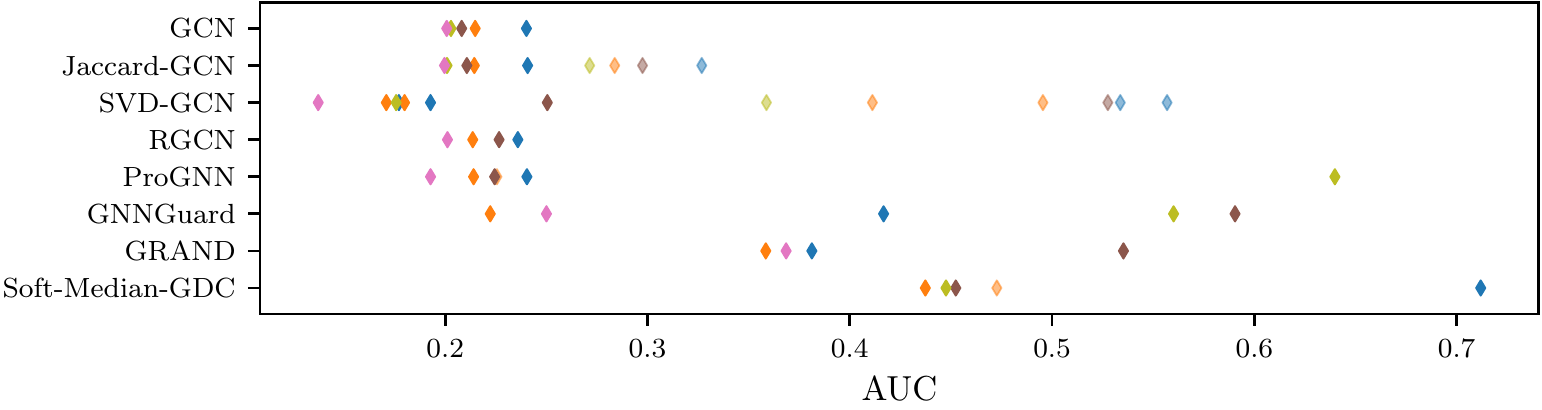}}

    \caption{The AUC of every local attack we have conducted. Attacks are color-coded by principal technique. Muted colors have the same signification as in \autoref{fig:appendix:global_method_comparison}. We observe that (1) greedy brute force is often the best attack, closely followed by PGD, while FGA is not as strong; (2) Nettack can rarely be made stronger by utilizing the target model's weights instead of a surrogate model (red); (3) many defenses successfully defend against Nettack; (4) against those defenses for which we have adapted Nettack, it becomes much stronger (muted vs. normal green); (5) the adaptions are also beneficial for other attacks, as those with muted colors are worse.}
    \label{fig:appendix:local_method_comparison}
\end{figure}

In \autoref{fig:appendix:global_envelope_support} we report which of the global attack techniques generate the strongest attacks, and in \autoref{fig:appendix:global_method_comparison}, we break down every global attack attempt. Analogously, in \autoref{fig:appendix:local_envelope_support} and \autoref{fig:appendix:local_method_comparison}, we report which local attack techniques require the smallest budget to misclassify the target nodes. In \autoref{fig:appendix:global_method_comparison}, we additionally compare different loss types for global attacks.

In general, we can say that PGD is the dominating attack for global evasion. For poisoning, Meta-PGD seems to be the strongest -- slightly more successful than Metattack, though not in every case. Greedy brute force dominates the local attacks, but for some defenses, PGD and Nettack have an edge.

\vspace{0.5cm}

\newpage
\section{Sensitivity to random seed}\label{sec:appendix:random_seed}

\begin{wrapfigure}[17]{r}{0.45\textwidth}
    \centering
    \vspace{-1.1cm}
    \includegraphics[width=\linewidth]{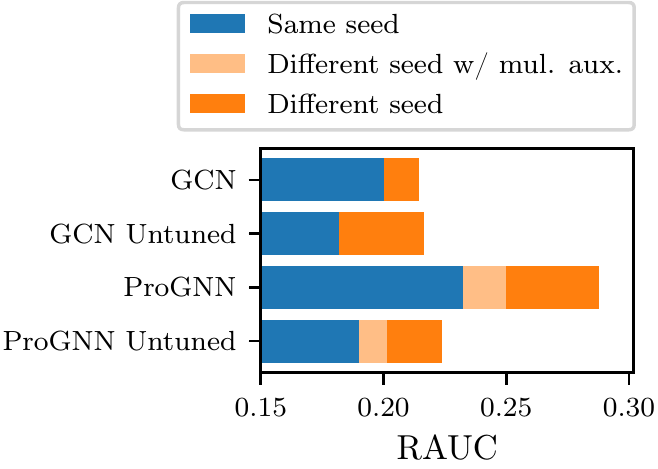}
    \caption{Lowest RAUC achieved by global evasion-poisoning transfer attacks on Cora ML under the premise that the random seed used by the victim is known respectively unknown to the attacker. While not knowing the seed is disadvantageous especially on ProGNN, our attack using multiple auxiliary models successfully compensates this issue.}
    \label{fig:appendix:random_seed_sensitivity}
\end{wrapfigure}

When transferring perturbations from evasion to poisoning, a different random seed is used for training the poisoned model than was used for the evasion one. In \autoref{fig:appendix:random_seed_sensitivity}, we study using the example of GCN and ProGNN whether poisoning success improves when we instead assume the same seed is used. This is indeed the case and turns out particularly strong on tuned ProGNN. However, by using multiple auxiliary models during evasion as detailed in \autoref{sec:methodology} under the ProGNN example subheading, we can substantially reduce the dependence of the attack upon a particular random seed and thereby improve attack performance.

\section{Robustness over node degree}\label{sec:appendix:node_degreee}

\begin{figure}[h]
    \centering

    \includegraphics[scale=0.78]{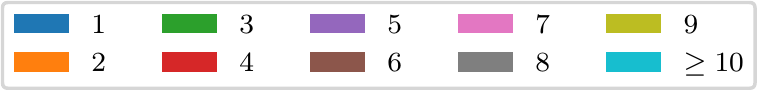}
    \hspace{2.1cm}
    \includegraphics[scale=0.78]{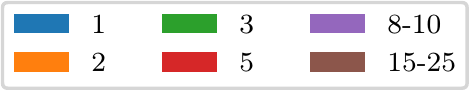}
    \hspace{0.5cm}

    \vspace{0.08cm}
    {
        \captionsetup[subfigure]{margin={0.7cm,0cm}}
        \subcaptionbox{Global, Cora~ML}{\includegraphics[scale=0.78]{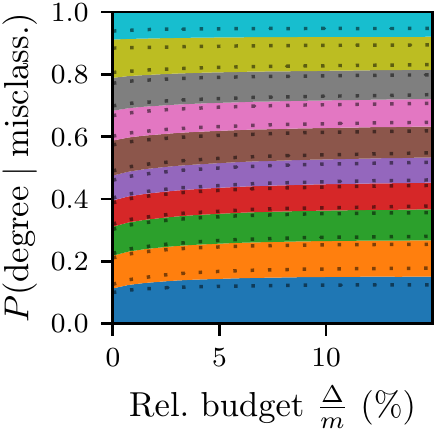}}
    }%
    \hspace{0.25cm}
    \subcaptionbox{Global, Citeseer}{\includegraphics[scale=0.78]{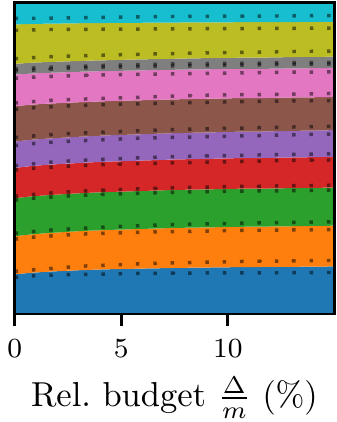}}
    \hfill
    {
        \captionsetup[subfigure]{margin={0.8cm,0cm}}
        \subcaptionbox{Local, Cora~ML}{\includegraphics[scale=0.78]{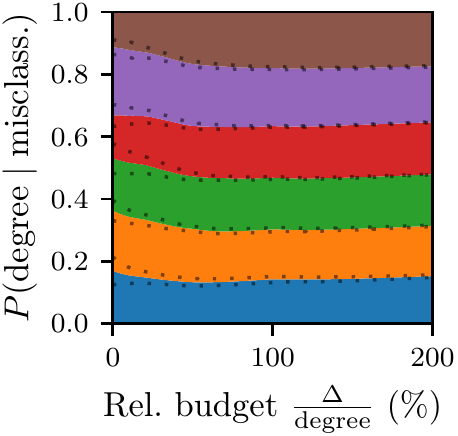}}
    }%
    \hspace{0.1cm}
    \subcaptionbox{Local, Citeseer}{\includegraphics[scale=0.78]{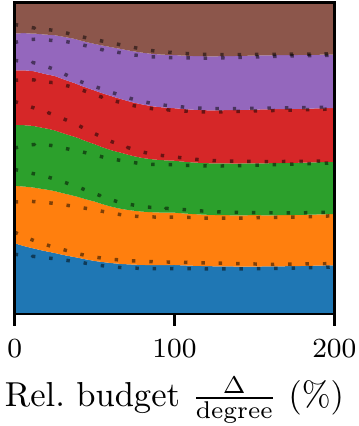}}

    \caption{The probability that a misclassified node is in a certain degree range. More specifically, for global attacks, that is which ratios of test set nodes from subsets with degree 1, 2, 3, ... , 9, \(\geq 10\) are misclassified per budget, normalized s.t. the stacked results sum to 1 everywhere. For local attacks, we show the amount of nodes from each target node set misclassified per budget, again normalized s.t. the stack sums to 1. Results are averaged over all experiments conducted (including evasion and poisoning) on tuned models. The dotted lines indicate standard deviation. We observe no substantial systematic bias towards the misclassification of low-degree nodes.}
    \label{fig:appendix:degree_preference_and_easiness}
\end{figure}

We explore the behavior of nodes under attack depending on their degree. In \autoref{fig:appendix:degree_preference_and_easiness}, we show the probability that a successfully misclassified node falls into a certain degree range, broken down by relativ budget.

We cannot confirm the prevalent conjecture that global attacks tend to target low-degree nodes, as they are easier to break. Our results show that all degree groups are misclassified uniformly over all budgets. There is no clear preference for lower-degree nodes.

For local attacks, on the other hand, we indeed observe that the success rate of changing the predicted class is independent of the node degree if and only if using a relative budget. For example, when allowing a certain relative budget, e.g., 100\% of the target node's degree, we manage to misclassify the same fraction of 1-degree target nodes (with absolute budget of 1) as 5-degree ones (with absolute budget of 5).

\vspace{0.3cm}

\newpage
\section{Attack characteristics}\label{sec:appendix:attack_characteristics}

\begin{wrapfigure}[37]{r}{0.52\textwidth}
    \vspace{-0.3cm}

    \captionsetup[subfigure]{skip=0.1cm}
    \subcaptionbox{Node degree}{\includegraphics[scale=0.7]{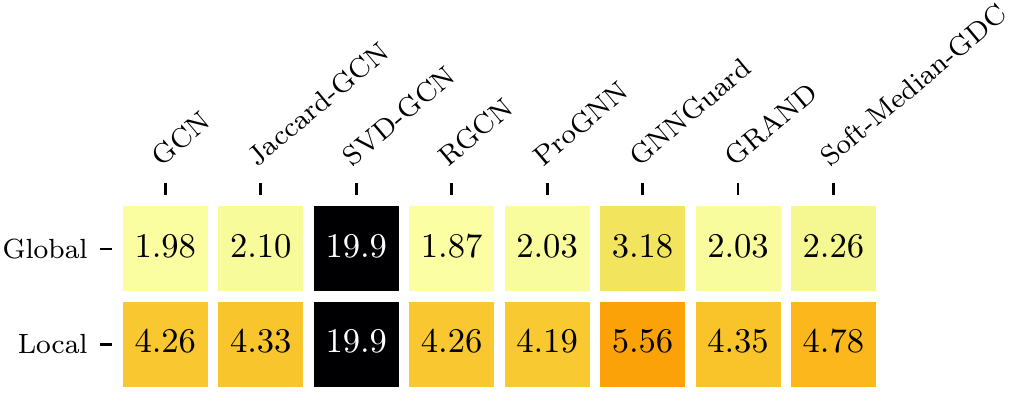}}

    \vspace{0.2cm}
    \captionsetup[subfigure]{margin={1cm,0cm}}
    \subcaptionbox{Closeness centrality}{\includegraphics[scale=0.7]{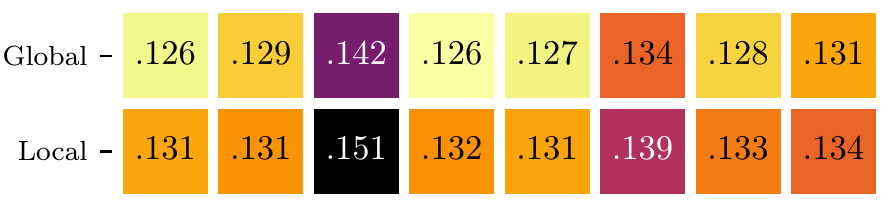}}

    \vspace{0.2cm}
    \subcaptionbox{Homophily}{\includegraphics[scale=0.7]{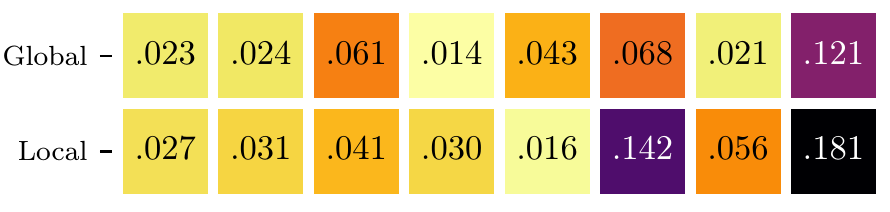}}

    \vspace{0.2cm}
    \subcaptionbox{Jaccard similarity}{\includegraphics[scale=0.7]{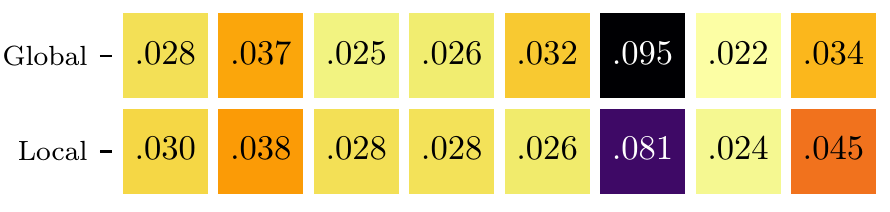}}

    \vspace{0.2cm}
    \subcaptionbox{Ratio of removed edges}{\includegraphics[scale=0.7]{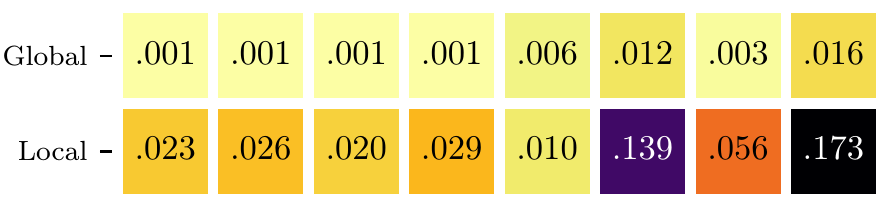}}

    \caption{Various metrics characterizing the nature of the adversarial edges from our strongest attacks, which are those visible in \autoref{fig:appendix:global_envelope_support} and \autoref{fig:appendix:local_envelope_support}, as well as the nature of the nodes connected respectively disconnected by them.}
    \label{fig:appendix:attack_char_matrix}
\end{wrapfigure}

Next, we present interesting patterns of the adversarial perturbations for each model/defense. We show the \emph{(1) node degree}, \emph{(2) closeness centrality}, \emph{(3) homophily}, \emph{(4) Jaccard similarity} of node attributes, and \emph{(5) the ratio of removed edges} over the strongest edge perturbations in \autoref{fig:appendix:attack_char_matrix}. For statistics 1-4, we consider the pairs of nodes that were affected by an adversarial edge flip (i.e., insertion or removal). Here we average over the strongest attack found for each budget (without transferring attacks between defenses). Thus, the values indicate what characteristics are important for strong, adaptive attacks.\looseness=-1

\textbf{(1) Node degree.}
For global attacks, the degree tends to be lower than the average degree of the dataset as given in \autoref{tab:appendix:datasets}. The higher average degree for local attacks might be influenced by the node selection. Interestingly, on SVD-GCN attacks connect very high-degree nodes, most likely because high-degree nodes correspond to dimensions represented by the most significant eigenvectors of \(\adj\) (see \autoref{sec:methodology} Example 1 and \autoref{sec:appendix:defenses_svd_gcn}). The attacks exploit the sensitivity of SVD-GCN to perturbations of high-degree nodes. This could hint towards how adaptive attacks catastrophically break SVD-GCN.

\textbf{(2) Closeness centrality.}
The closeness centrality of a particular node \(v\) is one over the sum of distances from \(v\) to all other nodes in the graph, multiplied by the total number of nodes in the graph. Attacks against SVD-GCN connect very central nodes, which probably correlates with them having high degrees. Interestingly, also the perturbations for GNNGuard seem to be of slightly increased centrality.

\textbf{(3) Homophily}
refers here to the fraction of pairs of nodes that share the same class. Successful adaptive attacks on Jaccard-GCN share the same homophily as those on GCN, indicating that the Jaccard coefficient is not suited to filter heterophil edges. Attacks on SVD-GCN, GNNGuard, and Soft-Median-GDC have higher homophily than those on GCN, hinting that these defenses successfully filter some heterogeneous edges, forcing some attacks to adapt.

\textbf{(4) Jaccard similarity.}
As expected, attacks on Jaccard-GCN have to compensate its filter by picking edges with nonzero coefficient. Attacks against GNNGuard connect nodes with very similar features, presumably to get past its cosine distance-based edge weighting. Curiously, attacks against Soft-Median-GDC behave similarly, yet only in the local setting and less pronounced. This is probably necessary to avoid that the new edges are weighted down as outliers by the robust aggregation, which becomes less of an issue when perturbing a large amount of edges in the global setting and thereby shifting what it means to be an outlier. Other defenses and especially GRAND admit connecting nodes as or more dissimilar than is the case on GCN.

\textbf{(5) Ratio of removed edges.}
It is clear to see that for all models, the adversarial attack mostly adds new edges. This indicates that edge insertion is stronger than edge deletion. Strong adaptive attacks on GNNGuard and Soft-Median-GDC seem to require the most edge deletions. Moreover, deletions are of much greater importance for local attacks.

\section{Spectral properties of adaptive attacks}\label{sec:appendix:spectral_attack_characteristics}

Previous studies have shown that adversarial attacks tend to focus the high-frequency (i.e., less significant) singular values of the adjacency matrix, both in the local~\citep{entezari_all_2020} and global~\citep{jin_graph_2020} setting. In consequence, defenses that exploit this observation to subdue attacks have been proposed (including SVD-GCN and ProGNN). This is a prime example of where (1) defenses were designed to circumvent specific attack characteristics and (2) an intuitive explanation exists of why the defense should improve robustness. However, our adaptive attacks have shown that neither (1) nor (2) entail actual robustness. In the case of SVD-GCN, it seems like the model becomes even less robust. It is only natural to ask whether our attacks exhibit spectral properties different from the high-frequency observation upon which SVD-GCN is built.

\begin{figure}[h]
    \centering

    \includegraphics[scale=0.8]{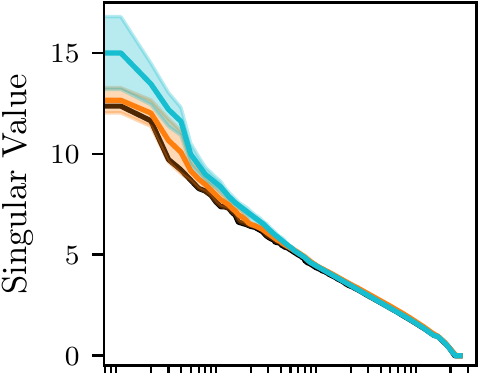}
    \hfill
    \includegraphics[scale=0.8]{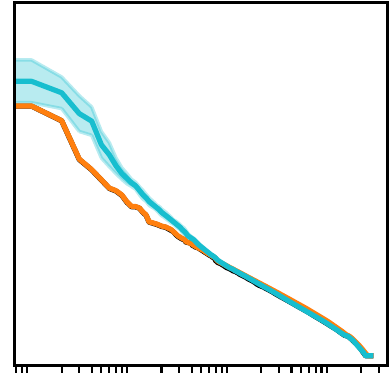}
    \hfill
    \includegraphics[scale=0.8]{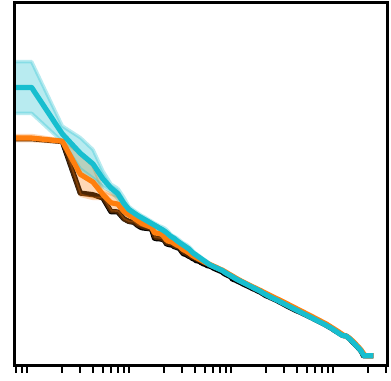}
    \hfill
    \includegraphics[scale=0.8]{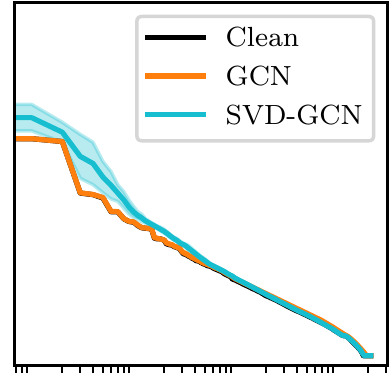}

    \vspace{0.15cm}

    {
        \captionsetup[subfigure]{oneside,margin={0.5cm,0cm}}
        \subcaptionbox{Cora ML, FGA/PGD}{\includegraphics[scale=0.8]{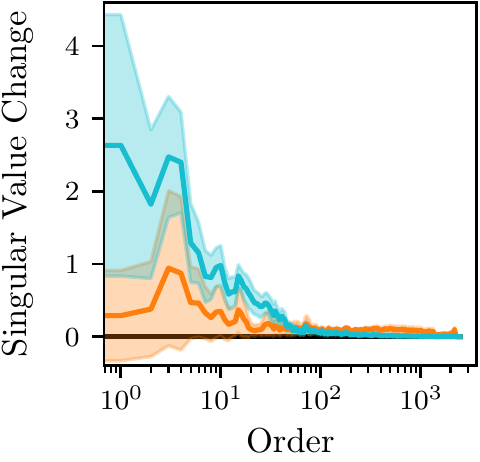}}
    }%
    \hfill
    \subcaptionbox{Cora ML, Meta}{\includegraphics[scale=0.8]{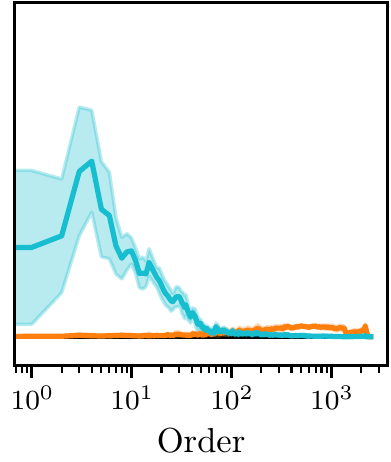}}
    \hfill
    \subcaptionbox{Citeseer, FGA/PGD}{\includegraphics[scale=0.8]{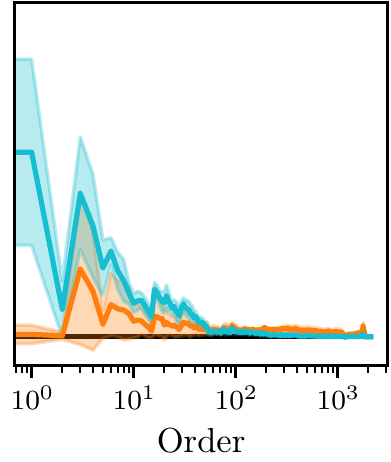}}
    \hfill
    \subcaptionbox{Citeseer, Meta}{\includegraphics[scale=0.8]{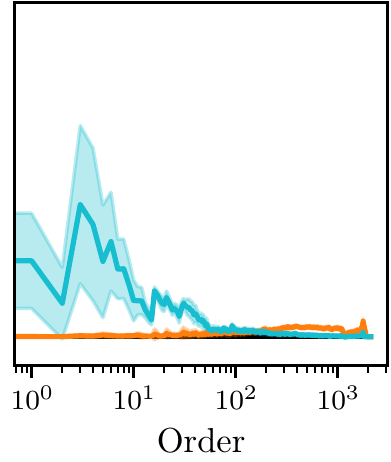}}

    \caption{Singular value spectra of the adjacency matrix before and after perturbation via global adaptive attacks with relative budget of 7.5\% against GCN and SVD-GCN. Results are split into native evasion attacks (via FGA and PGD) and native poisoning attacks (via Metattack and Meta-PGD), and averaged in each group. The top row shows the absolute spectrum, and the bottom row the difference to the clean spectrum. The order is plotted logarithmically. We observe that attacks against SVD-GCN strongly perturb the low-order singular values, and it is evident from the relative plots that high-order singular values are perturbed less compared to attacks against GCN.}
    \label{fig:appendix:spectral_attack_char}
\end{figure}

In \autoref{fig:appendix:spectral_attack_char}, we show the spectra of adjacency matrices before and after attacking GCN and SVD-GCN in various settings. Indeed, our adaptive attacks on SVD-GCN perturb more of the low frequencies and less of the high frequencies compared to attacks on GCN. Even though such low frequency-heavy perturbations are hypothesized to be ``noticeable''~\citep{entezari_all_2020,jin_graph_2020}, it is unclear how this can be exploited in practice without knowing the clean graph or the underlying distribution of the spectrum. In \autoref{sec:appendix:attack_overview}, we give additional reasons why we disregard constraints beyond the \(L_0\) difference.

\autoref{fig:appendix:spectral_attack_char} also shows that, in contrast to previous beliefs, effective attacks on a GCN may lie in the low-frequency spectrum (see subplots a and c). This questions the strategy of dampening high-frequency singular values to defend against attacks in the first place.

\vspace{0.5cm}

\newpage
\section{On the scalability of adaptive attacks}\label{sec:appendix:scalability}

In our main paper, we do not study adversarial robustness on larger graphs as (a) most defenses do not scale well and (b) we do not want to distract from our finding that structure defense evaluations are overly optimistic. Nevertheless, we consider scalability to be an important aspect for robustness as it is relevant for many applications. As mentioned in \autoref{sec:related_work}, \citet{geisler_robustness_2021} already study adaptive attacks scaled to large graphs. However, their work is focused on their own defense, and they only consider evasion. For these reasons, we now briefly discuss adaptive attacks on larger graphs.

\begin{figure}[h]
    \centering
    {
        \captionsetup[subfigure]{oneside,margin={0.7cm,0cm}}
        \subcaptionbox{Poisoning}{\includegraphics[scale=0.8]{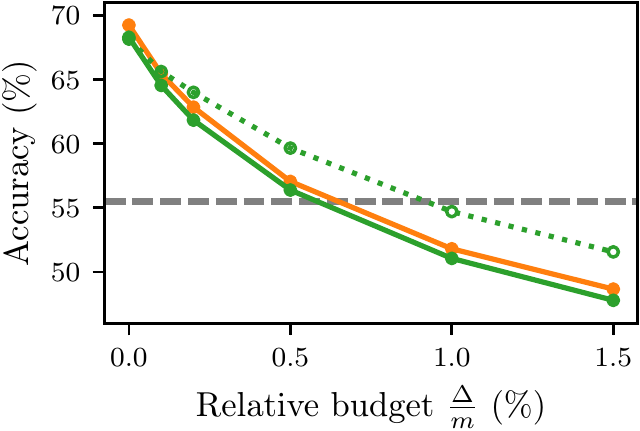}}
    }%
    \hspace{0.1cm}
    {
        \captionsetup[subfigure]{oneside,margin={-2.7cm,0cm}}
        \subcaptionbox{Evasion}{\includegraphics[scale=0.8]{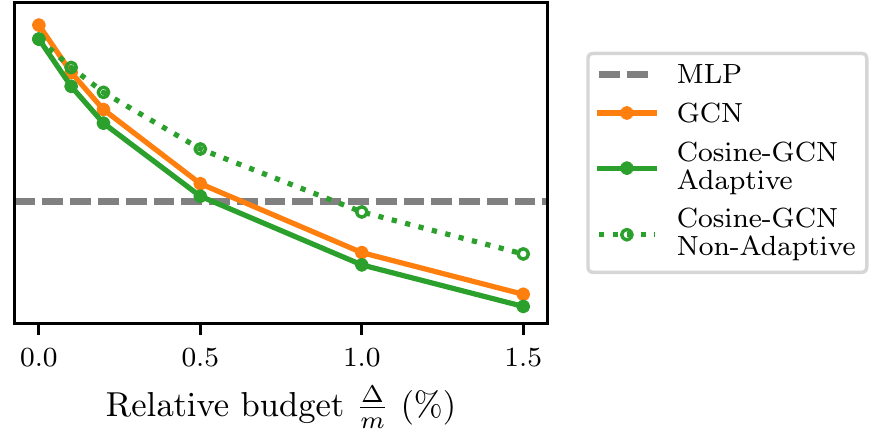}}
    }%
    \caption{Adversarial accuracy on the large arXiv dataset per budget for the scalable PRBCD attack against a regular GCN and our Cosine-GCN (single random seed). We use a block size of 1 million edges and run the attack for 200 epochs. Thereafter, we keep the best block for another 50 epochs fixed. Poisoning is conducted by transferring perturbations from evasion.}
    \label{fig:appendix:foreign_scalability_global_mr_curves}
\end{figure}

In \autoref{fig:appendix:foreign_scalability_global_mr_curves}, we show an adaptive attack against ``Cosine-GCN'' on arXiv from the Open Graph Benchmark~\citep{hu_open_2020} (169k nodes). Our Cosine-GCN defense is a natural equivalent of Jaccard-GCN~\citep{wu_adversarial_2019} for continuous features. Similarly to Jaccard-GCN on the smaller graphs, Cosine-GCN also comes with some robustness w.r.t. a non-adaptive attack. However, once we apply an adaptive attack, it performs actually slightly worse than the GCN baseline.

\textbf{Scaling first order attacks.}
The biggest challenge is certainly that the number of elements in the adjacency matrix scales quadratically with the number of nodes. One way to circumvent this ``curse of dimensionality'' is to use randomization. For our adaptive attack, we adopt Projected Randomized Block Coordinate Descent (PRBCD)~\citep{geisler_robustness_2021}. PRBCD uses the same relaxation as PGD (see \autoref{sec:preliminaries} and \autoref{sec:appendix:attack_overview}). In each iteration of the attack, it considers only a random subset of edges for gradient update and subsequent projection. Then, for the next iteration, PRBCD keeps edges of high weight and randomly re-samples the edges of low weight. This way, the overhead remains constant in the block size. Since PRBCD is a first-order attack, it is natively adaptive for differentiable models.

\textbf{Evasion vs. poisoning.}
Gradient-based poisoning attacks seem inherently more challenging since we need to unroll the training. Nevertheless, as long as we can run an evasion attack, there is the possibility to transfer the perturbed adjacency matrix to the poisoning setting. Here, we chose this approach. Still, \citet{zugner_adversarial_2019} show in their appendix that only very few training steps are actually required for Metattack to be effective. Using a low number of training steps is therefore something to consider to scale direct poisoning attacks on larger graphs.

\end{document}